# HIERARCHICAL MODELING
# OF MULTIDIMENSIONAL DATA
# IN REGULARLY DECOMPOSED SPACES
## TOME 4 : SYNTHESIS AND PERSPECTIVE
## (2016 – 2018)

- 2020 -

Olivier Guye

**Abstract of tome 4 :**

This fourth and last tome is focusing on describing the envisioned works for a project that has been presented in the preceding tome. It is about a new approach dedicated to the coding of still and moving pictures, trying to bridge the MPEG-4 and MPEG-7 standard bodies.

The aim of this project is to define the principles of self-descriptive video coding. In order to establish them, the document is composed in five chapters that describe the various envisioned techniques for developing such a new approach in visual coding:
- image segmentation,
- computation of visual descriptors,
- computation of perceptual groupings,
- building of visual dictionaries,
- picture and video coding.

Based on the techniques of multiresolution computing, it is proposed to develop an image segmentation made from piecewise regular components, to compute attributes on the frame and the rendering of so produced shapes, independently to the geometric transforms that can occur in the image plane, and to gather them into perceptual groupings so as to be able in performing recognition of partially hidden patterns.

Due to vector quantization of shapes frame and rendering, it will appear that simple shapes may be compared to a visual alphabet and that complex shapes then become words written using this alphabet and be recorded into a dictionary. With the help of a nearest neighbour scanning applied on the picture shapes, the self-descriptive coding will then generate a sentence made from words written using the simple shape alphabet.

**Fields :** Modeling and simulation, Algorithm and data structure, Image processing, Computer-based vision and pattern recognition, Learning

# Table of contents

# Introduction

This fourth and last tome is focusing on describing the envisioned works for a project that has been presented in the preceding tome. It is a project about a new approach dedicated to the coding of still and moving pictures, trying to bridge the MPEG-4 and MPEG-7 standard bodies. The aim of this project is to define the principles of self-descriptive video coding.

The envisioned paradigm relies on a layered representation model that gives structure to the information belonging to still and moving pictures, that is the information made from pixel rows, geometric structures, relations shared by regions and visual objects. These different kinds of information are complementary to each other and can be retrieved one from the another one according to a progressive approach naturally leading to a pyramidal organization of visual information.

The first aim is then to develop an intermediate level representation model that allows to handle the picture and video content starting from their homogeneous components. A geometrically representation is also defined for these various components.

These components are visual primitives established by learning from the captured content, then recorded into a dictionary. They can be managed independently from each other or gathered in order to make easier editing and content-based searching.

Shape descriptors are used for coding, recording and indexing various visual primitives and eventually the full content of still and moving pictures.

# 1. Description of the work

The work plan is divided into several steps enabling to develop the necessary technologies for achieving the aims of the research project.

It includes the following steps:

- image segmentation,
- computation of visual descriptors,
- computation of perceptual groupings,
- building of visual dictionaries,
- still and moving picture coding.

The first step will be focused on the image structure interpreted as piece-wise regular surfaces and will try to find the order of these surface pieces in distinguishing the singularities from the encountered irregularities. It will enable to develop an image segmentation, not driven by the search of adjacent points according to a metric topology, but rather by favoring the ultra-metric relations between regions coming from the piece-wise segmentation: the notion of visual object is then replaced by the viseme one which is a smaller homogeneous and compact unit.

The second step is looking for measuring these new visual shapes by using generalized moments, that enable to localize and to provide measures invariant to geometric transformations that the support of these simple shapes may suffer inside the image plane, and the reduction of the analytic expressions of the shape rendering in the tridimensional space linked to the corresponding surface piece, knowing that a single viseme will include as many tridimensional shapes as there are spectral bands in the image.

The following step will be interested in the compound shapes obtained by hierarchical aggregation of simple shapes in the image in the form of perceptual groupings ([15],[16]) that will be nearer to the classical notion of visual objects and that will enable to develop a new technique for shape recognition in hidden parts at halfway from the statistical ([9]) and structural ([10]) shape recognition methods. The computation of descriptors linked to simple shapes will be extended to complex shapes in order to provide a single way for calculating descriptors whatever is the nature of the shape.

Then our attention will be brought on the quantization of their descriptors that will be used as keys for retrieving the visual information from a data base and the synthesis techniques necessary for rebuilding an image with the help of shapes registered in the data base. It will appear that as soon as they are quantized the simple shapes make up a visual alphabet and that data bases in which complex shapes are registered are then dictionaries of words written with this visual alphabet.

At last in order to encode shapes belonging to an image, we will be interested in curves that fulfill plane and more especially in the Peano-Hilbert curve ([8]) that has the property to provide a nearest neighbor path in a space of any dimension. It will then appear that the corresponding path makes up a sentence from words belonging to the previous described dictionary and that its syntax may allow to define more globally the nature of a visual scene and the kind of picture scene that has been captured during shooting.

# 2. Image segmentation

## *2.1. Introduction*

The project is relying on the works led in the past in ADERSA in order to develop a new modeling technique based on a multiple piecewise regression for continuous process control ([28]).

It is a regression technique based on the recursive dividing of a data set applied in an orthogonal way to its main inertia axis using the hyperplane crossing the gravity center of its associated points cloud, until obtaining a minimal approximation error.

The result of this decomposition led in structuring data along a binary tree where data is gathered into subsets of neighboring data fitting a same linear model according to a given approximation error.

This approach is belonging to the field of classification and regression trees also named CARTs .

## *2.2. Experimental foundations of a regular piecewise decomposition*

### 2.2.1. Taking in account the irregularities of a model

This modeling process has been used for developing parametric command laws in the field of continuous process control.

In order to get the linearization of a higher order process, data is re-sampled by inserting as many delays on the output as necessary in order to be able to take in account the physical system order when this one is known. This allows to set up a finite difference scheme inside the initial data and to estimate the system dynamic behavior, for instance according to its speed and its acceleration.

In use this approximation method showed continuity issues when getting away from the gravity centers of data clouds for crossing from one subset to another neighbor one according to the recursive dividing of modeling data.

In order to reduce the generation of irregularities when using a piecewise linear model, it has been developed a continuous fitting technique relying on the barycentric interpolation of estimated data over two connected pieces inside the modeling tree at the neighborhood of their separating hyperplane.

### 2.2.2. Adjacent linear model fitting

The continuous fitting is computed as the composition of forms at the straight higher order according to the following way:

$$f = \frac{|\overrightarrow{C_g M} \cdot \overrightarrow{C_g C_d}|}{\|\overrightarrow{C_g C_d}\|^2} \cdot f_g + \frac{|\overrightarrow{MC_d} \cdot \overrightarrow{C_g C_d}|}{\|\overrightarrow{C_g C_d}\|^2} \cdot f_d$$

where $M$ is the point in the space where it is expected to know an estimated value,

$C_g$ and $C_d$ are the gravity centers of point clouds linked to the left and right children of a node in the modeling tree,

$f_g$ and $f_d$ are the regression forms computed over these two data clouds.

It is similar to a multidimensional fitting scheme of a B-Spline function where the support points are the gravity centers of the point clouds connected according to their separating hyperplane. It produces a new model of directly upper order to the two previous models, that are aggregated by regularizing the crossing from one to another one at the neighborhood of their common boundary, and this according to all the variables of the two initial models, this meaning in a multiple way.

If the modeling data are corresponding to infinitely continuous and differentiable forms the aggregation scheme can be applied in a recursive manner up to the root of the modeling tree : the number of levels of the tree will define the continuous model order so generated.

In practice, it has been implemented using a tuning parameter enabling to reduce its action to the neighborhood of the separating hyperplanes of the piecewise linear model and it has given users satisfaction for reducing bounces when applying control command laws.

### 2.2.3. Generation of continuous models of any order

The provision of a piecewise linear model regularized up to the order allowed by the hierarchical decomposition of a digitized data set, is monitored by an error threshold given at the generation of a linear model before its regularization. It corresponds to an approximation method trying to find a parametric model satisfying this threshold.

The local fit of a model over a digitized data set allows to know until which precision the data is following a linear continuous behavior by smoothing the noise belonging to the modeling data set.

Let us forget for a while the presence of this noise and let us put our attention on the only interpolation scheme provided by the B-Spline fitting.

By relying on the multidimensional data set hierarchically modeled in a space regularly decomposed ([28]), let us now directly apply this interpolating scheme on the data belonging to the nodes of the deepest level in the data set modeling tree and merge two by two nodes while climbing progressively the tree up to reach its root.

At the deepest level in the tree, these data are corresponding to a step function from which it will be possible to compute upper order interpolating functions by successively applying in each space variable the following scheme:

$$f = \frac{x_i - x_{g,i}}{x_{d,i} - x_{g,i}} \cdot f_g + \frac{x_{d,i} - x_i}{x_{d,i} - x_{g,i}} \cdot f_d$$

where $i$ is the index number of this space direction and $x_i$ the corresponding coordinate of point $M$,

$x_{g,i}$ and $x_{d,i}$ are the same index coordinates of the gravity centers of the point clouds linked to left and right children of a father node,

$f_g$ and $f_d$ are the forms recursively computed from the functionals recorded in terminal nodes with the help of this interpolation scheme.

By aggregating in such a way the forms met in climbing the tree of the interpolating pyramid, if the forms linked to two children nodes are the same the recursive interpolation scheme will generate the same form at father node level. It will be then locally reached the maximum interpolation order for this data set. If the step function corresponds to the digitization of a continuously differentiable function, we will get its maximum order polynomial expansion according to each variable of the data digitization coordinate system.

## *2.3. Piecewise regular decomposition of an image*

### 2.3.1. Digital representation of an image

A digital image can be represented in the form of one or several functions from $[0, x_M] \times [0, y_N] \subset \mathbb{R}^2$, domain of the $N$ rows and $M$ columns of the image, towards:

- $[0, z_G] \subset \mathbb{R}$, for a monochromatic image or the luminescence of a color image;

- $[0, z_{C_R}] \times [0, z_{C_B}] \subset \mathbb{R}^2$, for the red and blue chrominances of a color image;

- $[0,z_1]\times[0,z_2]\times\ldots\times[0,z_{N_s}]\subset\mathbb{R}^{N_s}$ , where $N^S$ is the number of spectral bands of a multispectral image.

Let $z=\begin{pmatrix} z_1 \\ z_2 \\ \vdots \\ z_{N_s} \end{pmatrix}$ , then the image digitization provides a set of discrete values over the image meshed domain:

- this one is planar and can be expressed $[0,N_C-1]\times[0,N_L-1]\subset\mathbb{N}^2$ where $N_L$ and $N_C$ are the image row and column numbers ;

- and on which are relying the values of the digital image, belonging to one of the possible spaces $[0,N_G-1]$, $[0,N_{C_R}-1]\times[0,N_{C_B}-1]$, $\prod_{k=1}^{N_S}[0,N_k-1]\subset\mathbb{N}^{N_S}$ where generally $N_G=N_{C_R}=N_{C_B}=N_1=\cdots=N_{N_S}=256$ .

For instance, here is shown hereinafter a color picture from which have been extracted on one hand the luminescence image, on the other hand the three colored components red, green and blue. The luminescence image is made from a single monochromatic image when the three colored component set constitutes a multi-chromatic image. These ones are displayed in the form of gray-level pictures, then as a surface drawing that each time described the patch of luminescent intensities in each bandwidth so as to well feel the globally continuous character of data at rough or mid-scale.

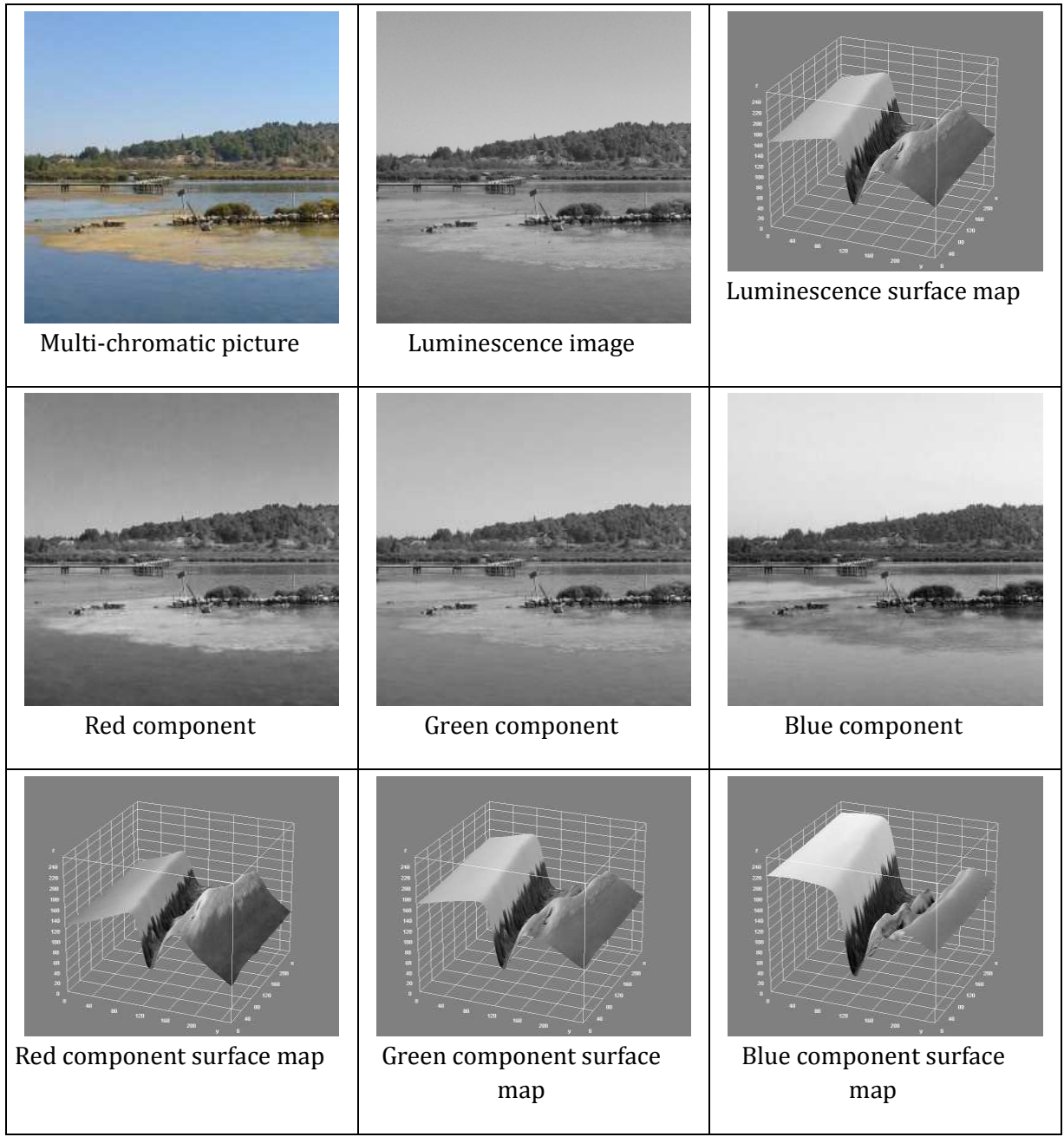

**Figure 1 : Digital picture representation**

### 2.5.2. Image least squares linear estimation

A digital image is then the following set of points :

$$\{((x_1, y_1), z_1); \cdots; ((x_{N_C \times N_L}, y_{N_C \times N_L}), z_{N_C \times N_L})\} = \{((x_k, y_k), z_k)\}$$

where $k = i + (j-1) \times N_C$, $i \in [1, N_C], j \in [1, N_L]$ and $z_k = \begin{pmatrix} z_{1,k} \\ z_{2,k} \\ \vdots \\ z_{N_S,k} \end{pmatrix} \in [0, N_G - 1]^{N_S}$

Then the image estimate using the least squares approach can be written as:

$$\hat{z} = \begin{pmatrix} f_1 \\ \vdots \\ f_l \\ \vdots \\ f_{N_S} \end{pmatrix} \begin{pmatrix} 1 \\ x \\ y \end{pmatrix} \quad \text{where } f_l \text{ are row vectors such as :}$$

$$\hat{z} = R_{z\binom{x}{y}} \cdot R^{-1}_{\binom{x}{y}\binom{x}{y}} \cdot \left( \binom{x}{y} - M\binom{x}{y} \right) + M_z$$

where $M\binom{x}{y} = \frac{1}{Card(V)} \begin{pmatrix} \sum_{p \in V} x_p \\ \sum_{p \in V} y_p \end{pmatrix} = \begin{pmatrix} \bar{x} \\ \bar{y} \end{pmatrix}$ and $M_z = \frac{1}{Card(V)} \begin{pmatrix} \sum_{p \in V} z_{1,p} \\ \sum_{p \in V} z_{2,p} \\ \vdots \\ \sum_{p \in V} z_{N_S,p} \end{pmatrix} = \begin{pmatrix} \bar{z}_1 \\ \bar{z}_2 \\ \vdots \\ \bar{z}_{N_S} \end{pmatrix}$

and $p = \binom{x}{y} \in V$ is a point from a subset belonging to the image domain,

where $R_{z\binom{x}{y}} = \frac{1}{Card(V)} \begin{pmatrix} \sum_{p \in V} (z_{1,p} - \bar{z}_1) \cdot (x_p - \bar{x}) & \sum_{p \in V} (z_{1,p} - \bar{z}_1) \cdot (y_p - \bar{y}) \\ \vdots & \vdots \\ \sum_{p \in V} (z_{N_S,p} - \bar{z}_{N_S,p}) \cdot (x_p - \bar{x}) & \sum_{p \in V} (z_{N_S,p} - \bar{z}_{N_S,p}) \cdot (y_p - \bar{y}) \end{pmatrix}$

and $R_{\binom{x}{y}\binom{x}{y}} = \frac{1}{Card(V)} \begin{pmatrix} \sum_{p \in V}(x_p - \bar{x})^2 & \sum_{p \in V}(x_p - \bar{x}) \cdot (y_p - \bar{y}) \\ \sum_{p \in V}(y_p - \bar{y}) \cdot (x_p - \bar{x}) & \sum_{p \in V}(y_p - \bar{y})^2 \end{pmatrix}$

As $R_{\binom{x}{y}\binom{x}{y}}$ is a symmetric positive definite matrix, it is then invertible and can be decomposed into singular values in the following way:

$$R_{\binom{x}{y}\binom{x}{y}} = U \Lambda U^{-1}, \text{ where } U^{-1} = U^T,$$

that is meaning where $\Lambda$ is the matrix of eigenvalues sorted in a decreasing way and $U$ is the matrix of eigenvectors of $R_{\binom{x}{y}\binom{x}{y}}$.

Knowing that the image is relying on a planar domain, it can be once more written:

$$R_{\binom{x}{y}\binom{x}{y}} = \begin{pmatrix} r_{xx} & r_{xy} \\ r_{yx} & r_{yy} \end{pmatrix} = \begin{pmatrix} \cos(\theta) & \sin(\theta) \\ -\sin(\theta) & \cos(\theta) \end{pmatrix} \begin{pmatrix} \lambda_1 & 0 \\ 0 & \lambda_2 \end{pmatrix} \begin{pmatrix} \cos(\theta) & -\sin(\theta) \\ \sin(\theta) & \cos(\theta) \end{pmatrix}$$

with $r_{yx} = r_{xy}$ and where

$$\lambda_1 = \frac{1}{2}\left(r_{xx} + r_{yy} + \sqrt{(r_{xx} - r_{yy})^2 + 4r_{xy}^2}\right),$$

$$\lambda_2 = \frac{1}{2}\left(r_{xx} + r_{yy} - \sqrt{(r_{xx} - r_{yy})^2 + 4r_{xy}^2}\right),$$

$$\theta = \arctan\left(\frac{\lambda_1 - r_{xx}}{r_{xy}}\right) \text{ more or less } \pi,$$

because $\lambda_1$ and $\lambda_2$ are the roots of the characteristic polynomial of $R_{\binom{x}{y}\binom{x}{y}}$, in other words the solutions of the equation $\lambda^2 - Tr(R_{\binom{x}{y}\binom{x}{y}}) \cdot \lambda + Det(R_{\binom{x}{y}\binom{x}{y}}) = 0$ and $U$ a matrix of rotation of angle $\theta$.

In order to get closer to the result of the application of a linear approximation in the least squares approach for an image using multiple resolution levels, it has been chosen to show it using a multi-resolution averaging of this same image, but restricted to the single luminescence image. At each different resolution level is drawn the corresponding map in addition to the averaged image.

| | | |
|---|---|---|
| <br>8x8 multi-chromatic | <br>8x8 mono-chromatic | 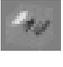 |
| <br>16x16 multi-chromatic | <br>16x16 mono-chromatic | 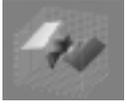 |
| 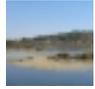<br>32x32 multi-chromatic | 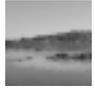<br>32x32 mono-chromatic | 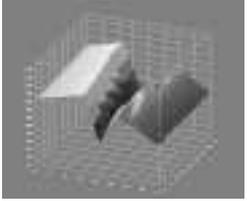 |
| 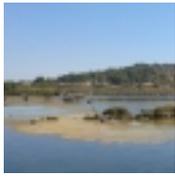<br>64x64 multi-chromatic | 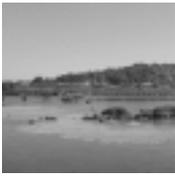<br>64x64 mono-chromatic | 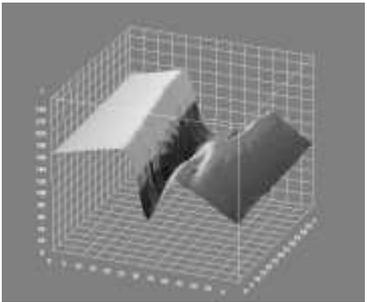 |
| 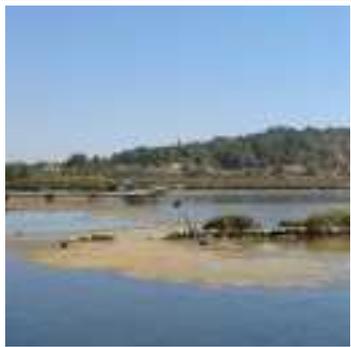<br>128x128 multi-chromatic | 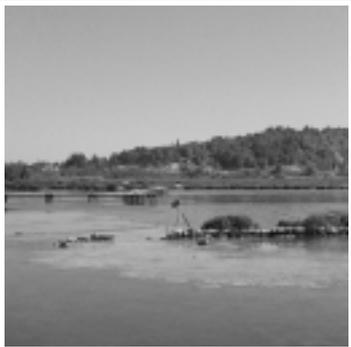<br>128x128 mono-chromatic | 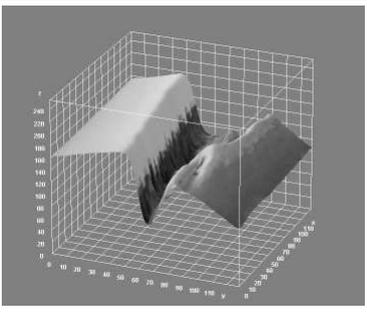<br>Corresponding surface maps |

**Figure 2 : Multi-resolution image approximation**

### 2.3.3. Irregularities detection and data set dividing

The aim is to localize the irregularities of present order and to extract a separating line inside the image planar domain enabling to divide the present data set into two more regular subsets.

Starting from the hypothesis that the singularities are the data that are at the source of the highest approximation errors, these ones are computed at each point of the data set :

$$\varepsilon_p^\infty = \lfloor Max\{|z_{1,p} - f_1(x_p, y_p)|, \cdots, |z_{N_s,p} - f_{N_s}(x_p, y_p)|\} \rfloor .$$

Then the histogram of approximation errors is built on the interval of integer values of the image co-domain: $[0, 2^{N_G} - 1]$

After that a threshold is determined so as to separate the singularities from the other approximation errors in the following way:

- either the histogram is mono-modal and the singularities are hidden in the approximation errors, then the threshold is placed before the $\sqrt{Card(V)}$ highest errors in order to be able to fit a regression straight line;

- or the histogram is bi- or multi-modal, then the threshold will be defined by the highest index valley so as to distinguish the singularities from the other errors that will be progressively reduced.

Let $V_S$ the subset of $V$ where the approximation errors are the most important, note by:

- $\bar{x}_S = \frac{1}{Card(V_S)} \sum_{p \in V} x_p$ , $\bar{y}_S = \frac{1}{Card(V_S)} \sum_{p \in V} y_p$ the center coordinates of $V_S$,

- $\sigma_{x^2} = \frac{1}{Card(V_S)} \sum_{p \in V} (x_p - \bar{x}_S)^2$ , $\sigma_{xy} = \frac{1}{Card(V_S)} \sum_{p \in V} (x_p - \bar{x}) \cdot (y_p - \bar{y}_S)$ the associated correlations.

Then the regression straight line fitted over the location of singularities has got as equation inside the image planar domain $y = \frac{\sigma_{xy}}{\sigma_{x^2}} \cdot (x - \bar{x}_S) + \bar{y}_S$, that is once more $(x - \bar{x}_S) \cdot \sigma_{xy} - (y - \bar{y}_S) \cdot \sigma_{x^2} = 0$ .

With the help of this equation, the $V$ domain can be divided into two subsets made from more regular data than that of the initial set by evaluating the sign of the following expression for each of its points:

- the points giving a negative value to this expression will be placed in $V_{-} = \{ p \in V / (x - \bar{x}_S) \cdot \sigma_{xy} - (y - \bar{y}_S) \cdot \sigma_{x^2} < 0 \}$ ;

- the ones providing a null or positive value to this expression will be placed in $V_{+} = \{ p \in V / (x - \bar{x}_S) \cdot \sigma_{xy} - (y - \bar{y}_S) \cdot \sigma_{x^2} \geq 0 \}$ ;

- at the end $V_{-} \cup V_{+} = V$ and $V_{-} \cap V_{+} = \emptyset$ .

### 2.3.4. Image piecewise linear decomposition

With the help of the previous results, a recursive process of image dichotomous decomposition can be performed in the following way:

- initially, the data set to process is composed of all the image data, that is meaning of $\{((x_k, y_k), z_k)\}$ where $k = i + (j-1) \times N_C$ with $i \in [1, N_C], j \in [1, N_L]$ and
$$z_k = \begin{pmatrix} z_{1,k} \\ z_{2,k} \\ \vdots \\ z_{N_S,k} \end{pmatrix} \in [0, N_G - 1]^{N_S} \; ;$$

- the least squares linear estimation of the data set can be written $\hat{z} = \begin{pmatrix} f_1 \\ \vdots \\ f_l \\ \vdots \\ f_{N_S} \end{pmatrix} \begin{pmatrix} 1 \\ x \\ y \end{pmatrix}$

  where the $f_l$ are the row vectors $f_l = (a_{1,l}, a_{x,l}, a_{y,l})$ and where $\hat{z}_l = a_{1,l} + a_{x,l} \cdot x + a_{y,l} \cdot y$ is the estimation of the $l^{i\text{ème}}$ image spectral band ;

- the possible irregularities present in the building data set can be detected in the following manner :

  - by building the histogram of approximation errors $H_\infty(V) = \{ Card(\{ p \in V / \epsilon_p^\infty = l \}), l \in [0, N_G - 1] \}$ where $\varepsilon_p^\infty = Max_{l \in [1, N_G]} \{ || z_{k,p} - f_k(x_p, y_p) || \}$ ,

- by selecting in $V$, the $V_S$ subset of points $p$ whose error $\varepsilon_p^\infty$ is higher than a threshold evaluated on the histogram $H_\infty(V)$ as previously quoted, which is $V_S = \{p \in V / \epsilon_p^\infty \geq seuil(H_\infty(V))\}$,

- then by computing inside the image planar domain the regression straight line of the points belonging to $V_S$, that is the equation of the straight line: $(x - \bar{x}_S) \cdot \sigma_{xy} - (y - \bar{y}_S) \cdot \sigma_{x^2} = 0$ where $\bar{x}_S$ and $\bar{y}_S$ are the coordinate averages of the points $p \in V_S$ and where $\sigma_{x^2}$ and $\sigma_{xy}$ are their variance and covariance,

- at last the analytical expression of this straight line is used for dividing the initial set $V$ into two subsets $V_-$ and $V_+$ such as: $V_- = \{p \in V / a_S \cdot x + b_S \cdot y + c_S < 0\}$ and $V_+ = \{p \in V / a_S \cdot x + b_S \cdot y + c_S \geq 0\}$ where $a_S = \sigma_{xy}$, $b_S = -\sigma_{x^2}$ and $c_S = \bar{y}_S \cdot \sigma_{x^2} - \bar{x}_S \cdot \sigma_{xy}$.

Due to this process, a set $V$ has been decomposed into two subsets $V_-$ and $V_+$ so as to gather closely to their boundary the strongest irregularities for reducing their number in each new subset.

The dividing process can be repeated on each subset provided by the decomposition until succeeding in globally fulfilling the same approximation error over the initial image set : it will be the fitting precision of the so produced image piece-wise linear model.

In order to perceive the kind of results that would provide such an approach, the previous multi-resolution display of the different averaging levels of an image is now complemented with the images of the differences between the average of a given level and its value before averaging. Then these difference images are thresholded according to a similar mechanism with the one that has been described in this paragraph so as to only keep the irregularities observed at the current level and drawn them as a surface form. At nearby a scale factor, the irregular areas can be retrieved from one level to another one, that should enable to progressively settle the separating straight lines over the ridges of these ones by successively analyzing every level until reaching the full resolution of the initial image. It would correspond to move from a conventional wavelet scheme to a geometrical wavelet scheme by trying to localize the areas where the image gets a regular behavior so as to be able to fit on them finite functional expansions.

| | | |
|---|---|---|
|  8x8 differences |  8x8 thresholding | 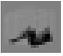 |
|  16x16 differences |  16x16 thresholding | 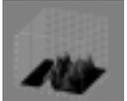 |
| 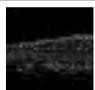 32x32 differences | 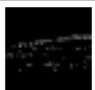 32x32 thresholding | 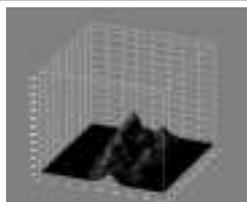 |
| 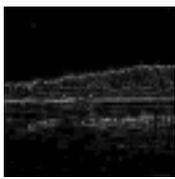 64x64 differences | 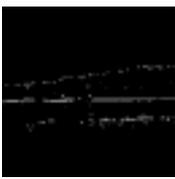 64x64 thresholding | 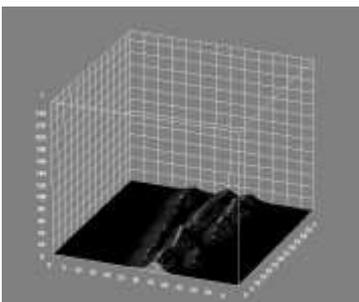 |
| 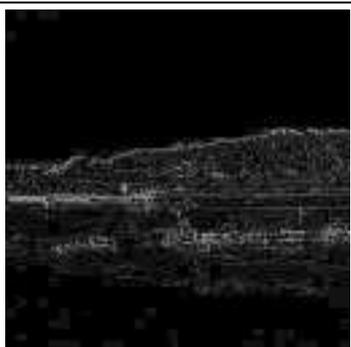 128x128 differences | 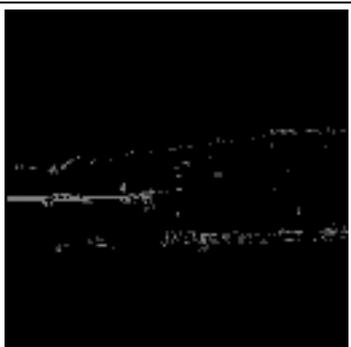 128x128 thresholding | 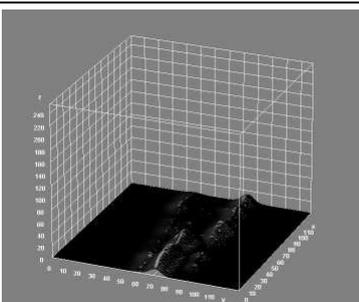 Corresponding surface maps |

**Figure 3 : Irregularities detection in an image**

### 2.3.5. Two image pieces aggregating at the next upper order

By starting this iterative process from a full image, it will be got a piecewise linear hierarchical decomposition in the form of a binary tree of first order continuously differentiable subsets.

For distinguishing upper order continuities from singularities found in the image, the uncertainty concerning the origin of these irregularities can be removed by successively aggregating models of the tree structure into next upper order models by climbing up the tree. By applying this principle to the two models linked to two terminal nodes children of a same tree node by combining them into a new upper order model that can be associated to their father node. It is necessary to assess the approximation error of the new model over the union of all the data belonging to the two pieces associated to the nodes to be aggregated. If the associated model error fulfills the expected fitting error, then it was truly a higher order irregularity and not a singularity locally dividing the data into two subsets : it can be envisioned to replace the two children nodes by the father node to which will be linked the new model coming from the aggregating of the two previous models. If it is not the case, then the aggregating process must be stopped and the two children nodes must be kept unchanged with their own models.

There is a third eventuality after having started to aggregate models: it may be that each model linked to two brother nodes can also apply themselves to the union of the data associated to these two nodes : it means that the model order do not temporally increase and that the two children nodes are equivalent at the expected fitting precision. The two children node models can be aggregated into a single one by associating to him the model of the node giving the best fitting error over the two node sets or also the average of the two children models. The phenomena may occur several successive times while climbing up to the root in the modeling tree.

The idea of aggregating two pieces of a given order in order to provide a new one at the consecutive upper order over the union of two adjacent domains is relying on the example of Bernstein polynomials that allow to progressively set up a polynomial interpolation scheme over a mesh of points representing a curve from which it is expected to get an analytic expression ([7]).

The scheme is the following one for a curve described by $n+1$ points $P_0, \ldots, P_n \in \mathbb{R}^m$,

$$m \geqslant 2 : \forall i \in [0,n], \ \forall t \in [0,1], \ B(t) = \sum_{k=0}^{n} P_k^0 B_k^n(t) = \sum_{k=0}^{n-i} P_k^i B_k^{n-i}(t) \text{, where the points } P_i^j$$

are recursively defined as the barycentric means $P_i^j = (1-t) P_i^{j-1} + t P_{i+1}^{j-1}$ and by applying the property of Bernstein polynomials $B_i^n(t) = (1-t) B_i^{n-1} + t B_{i+1}^{n-1}$ .

This scheme is at the foundation of the building of Bézier curves and their composite version the B-splines.

Due to the creation of intermediate points by barycentric averaging, the Bézier polynomials are built over a succession of convex combinations which implies that the corresponding curve stands inside the convex hull of the points of this curve carrier. This enables to tune the variations of such a polynomial curve and provides it a good regularity.

Concerning surfaces, it must be introduced the notion of bi-parametric tile $0 \leqslant u \leqslant 1$ and $0 \leqslant v \leqslant 1$ where a surface stands in a tridimensional space and be applied the rules of tensor products : $S(u,v) = \sum_{i=0}^{n} \sum_{j=0}^{n} P_{i,j} B_i^n(u) B_j^n(v)$ for a polynomial interpolation of order $n$ in each variable of the tile.

The proposed way for performing the model aggregation in the aim to get an upper order model takes the use of barycentric averaging by resting on the gravity centers of every data subsets coming from the image hierarchical decomposition. It then leads towards a polynomial approximation of order $n$ no more in each variable of the surface manifold domain, but in all the variables of this one.

By taking back the notations used for describing the dividing process of a set $V$ into two subsets $V_-$ and $V_+$, let us call:

- $M_-$ and $M_+$ the gravity centers such as they have been computed during the computing of the image piecewise linear estimation according to least squares,

- $\widehat{z_-}$ and $\widehat{z_+}$ the corresponding estimates on these two subsets,

then the barycentric interpolation scheme over these two subsets $V_-$ and $V_+$ generates the straight upper order estimate in all the variables as it follows:

$$\hat{z} = \frac{\langle \overrightarrow{M_- M} \cdot \overrightarrow{M_- M_+} \rangle}{\| \overrightarrow{M_- M_+} \|^2} \cdot \widehat{z_-} + \frac{\langle \overrightarrow{MM_+} \cdot \overrightarrow{M_- M_+} \rangle}{\| \overrightarrow{M_- M_+} \|^2} \cdot \widehat{z_+} \quad .$$

That is by developing in a matrix manner :

$$\begin{pmatrix} f_1 \\ \vdots \\ f_l \\ \vdots \\ f_{N_s} \end{pmatrix} = \frac{(x-\overline{x_-})\cdot(\overline{x_+}-\overline{x_-})+(y-\overline{y_-})\cdot(\overline{y_+}-\overline{y_-})}{(\overline{x_+}-\overline{x_-})^2+(\overline{y_+}-\overline{y_-})^2} \cdot \begin{pmatrix} f_1^- \\ \vdots \\ f_l^- \\ \vdots \\ f_{N_s}^- \end{pmatrix} + \frac{(\overline{x_+}-x)\cdot(\overline{x_+}-\overline{x_-})+(\overline{y_+}-y)\cdot(\overline{y_+}-\overline{y_-})}{(\overline{x_+}-\overline{x_-})^2+(\overline{y_+}-\overline{y_-})^2} \cdot \begin{pmatrix} f_1^+ \\ \vdots \\ f_l^+ \\ \vdots \\ f_{N_s}^+ \end{pmatrix}$$

where M is the point with coordinates $\begin{pmatrix} x \\ y \end{pmatrix} \in V = V_- \cup V_+$ .

As in the building scheme of Bézier polynomials, the process of model aggregation can be recursively applied by climbing the model tree up to stop when the approximation error is no more controlled. If the process can carry on up to the tree root then the image is only made from a single surface manifold which differentiability and continuousness order is equal to the modeling tree depth before anyone aggregation operation.

The first aggregation pass will generate the expression of a quadric, that is a polynomial of order 2 in all the variables $x$ and $y$ of the domain of each functional $f_l$. The second aggregation pass the expression of a cubic, a polynomial of order 3 for every same functionals. In order to provide rendering descriptors by using these pieces of information, it will be not necessary to go beyond the order 3, as it will be shown afterwards.

Working with locally continuous linear estimates allows to free ourselves in some way from the influence of capture and digitization noise and not to have to preprocess data before its use. The control is performed by the expected fitting precision of the estimate mean quadratic error over its domain.

The hierarchical aggregation using increasing order polynomials does not enable to provide an image segmentation according to the neighborhood systems classically associated to metric distances, but rather an image partition according to the ultra-metric distance induced by the image treelike division. It is a topology rougher than a metric topology that provides a more broken up segmentation than a classical segmentation. This characteristic should not hamper the processes of image coding and pattern recognition.

At the opposite the predicate used for segmenting images is not an iso-color predicate, but an iso-model one.

In place of a model barycentric combination, another method can be also envisioned for replacing the aggregation of two tree brother models into a new upper order model, it is the linearization of the problem at the order just upper from the initial models and by finding it using the least squares approach , that is to find:

- the estimate of order 2 in all the variables that minimize the mean quadratic error on the new aggregated data set if the initial models are linear, being the quadrics

  $$\hat{z}_l = a_{1,l} + a_{x,l} \cdot x + a_{y,l} \cdot y + a_{x^2,l} \cdot x^2 + a_{xy,l} \cdot xy + a_{y^2,l} \cdot y^2 \ ,$$

- then at next hierarchical level, the cubics,

$$\hat{z}_l = a_{1,l} + a_{x,l} \cdot x + a_{y,l} \cdot y + a_{x^2,l} \cdot x^2 + a_{xy,l} \cdot xy + a_{y^2,l} \cdot y^2 + a_{x^3,l} \cdot x^3 + a_{x^2y,l} \cdot x^2y + a_{xy^2,l} \cdot xy^2 + a_{y^3,l} \cdot y^3 ,$$

- etc.

If the linearized expressions are similar to those provided by barycentric combination, then it can be considered that the aggregation by a barycentric combination is the fast transform of increasing order linearized models. As the two methods provide finite expansions at a given order in all the variables, it can be enable to rely on the properties of Taylor expansions for insuring the uniqueness of these expansions whatever are their orders, when they are existing ([6]).

Whatever is the used approach, it should be better to priory check if lower order models can directly apply on the union of the two domains while fulfilling the expected modeling precision. If it is the case the best lower order model is propagated up to the tree higher level. If it can be possible to aggregate children models by climbing up in the tree, it means that the maximum modeling order has been reached.

After aggregating models up to the $p$ order, as well as their associated tree nodes, the set of resulting parts constitute an image partition into data subsets that can be approached by piecewise models of $p$ order. Note $\{V_i\}_{i \in 1, n(p)}$ the image partition where $n(p)$ is the number of class $C^p$ parts representing the initial image modeled at the expected precision.

## 2.4. Main algorithm for image any order regular decomposition

On the basis of the previous developments, it can be build an algorithm that should perform an image piecewise regular decomposition of any order:

BEGIN

Starting from all the image points is constituted the data set:

$V = \{((x_k, y_k), z_k)\}$ where $k = i + (j-1) \times N_C$

with $i \in [1, N_C], j \in [1, N_L]$ and $z_k = \begin{pmatrix} z_{1,k} \\ z_{2,k} \\ \vdots \\ z_{N_s,k} \end{pmatrix} \in [0, N_G - 1]^{N_s}$

order $\leftarrow 1$, $\hat{z} \leftarrow$ Piecewise regular decomposition ($V$, order, precision)

END

FUNCTION Piecewise regular decomposition ( $V$ , order, precision)

BEGIN

    Computation of the linear estimate of $V$ according to the least squares method :

$$\hat{z} = \begin{pmatrix} f_1 \\ \vdots \\ f_l \\ \vdots \\ f_{N_s} \end{pmatrix} \begin{pmatrix} 1 \\ x \\ y \end{pmatrix} \quad \text{where the } f_l \text{ are row vectors such as :}$$

$$\hat{z} = R_{z\binom{x}{y}} \cdot R^{-1}_{\binom{x}{y}\binom{x}{y}} \cdot \left( \binom{x}{y} - M\binom{x}{y} \right) + M_z \text{, where } \theta = \arctan\left(\frac{\lambda_1 - r_{xx}}{r_{xy}}\right)$$

and $\quad \lambda_1 = \frac{1}{2}(r_{xx} + r_{yy} + \sqrt{(r_{xx} - r_{yy})^2 + 4 r_{xy}^2})$ , $\lambda_2 = \frac{1}{2}(r_{xx} + r_{yy} - \sqrt{(r_{xx} - r_{yy})^2 + 4 r_{xy}^2})$

    Computation of the maximum approximation error:

$$\varepsilon_V^\infty = \lfloor Max_{p \in V} \{|z_{1,p} - f_1(x_p, y_p)|, \cdots, |z_{N_s, p} - f_{N_s}(x_p, y_p)|\} \rfloor$$

IF ( $\varepsilon_V^\infty$ > precision) THEN DO

    Computation of the error histogram :

$$H_\infty(V) = \{Card\,(p \in V / \epsilon_p^\infty = l), l \in [0, N_G - 1]\}$$

    IF ( mono-modal histogram)

    THEN $V_S \leftarrow \{ \sqrt{Card\,(V)} $ points of stronger error $\}$

    ELSE DO

        Define as threshold the histogram highest valley

$$V_S \leftarrow \{p \in V / \epsilon_p^\infty > threshold\}$$

    END

Compute the coordinates $\bar{x}$ and $\bar{y}$ of the gravity center

and the variance $\sigma_{x^2}$ and covariance $\sigma_{xy}$ of $V_S$

Divide the set $V$ into two subsets :

$$V_- = \{p \in V / (x-\bar{x}) \cdot \sigma_{xy} - (y-\bar{y}) \cdot \sigma_{x^2} < 0\}$$

$$V_+ = \{p \in V / (x-\bar{x}) \cdot \sigma_{xy} - (y-\bar{y}) \cdot \sigma_{x^2} \geq 0\}$$

$\widehat{z_-} \leftarrow$ Piecewise regular decomposition( $V_-$ , order, precision)

$\widehat{z_+} \leftarrow$ Piecewise regular decomposition ( $V_+$ , order, precision)

IF (order < 3) THEN DO

    Computation of the barycentric combination :

$$\hat{z} = \frac{(x-\overline{x_-}) \cdot (\overline{x_+}-\overline{x_-}) + (y-\overline{y_-}) \cdot (\overline{y_+}-\overline{y_-})}{(\overline{x_+}-\overline{x_-})^2 + (\overline{y_+}-\overline{y_-})^2} \cdot \widehat{z_-} + \frac{(\overline{x_+}-x) \cdot (\overline{x_+}-\overline{x_-}) + (\overline{y_+}-y) \cdot (\overline{y_+}-\overline{y_-})}{(\overline{x_+}-\overline{x_-})^2 + (\overline{y_+}-\overline{y_-})^2} \cdot \widehat{z_+}$$

    Computation of the approximation error :

$$\varepsilon_V^\infty = \lfloor Max_{p \in V} \{|z_{1,p} - f_1(x_p, y_p)|, \cdots, |z_{N_s, p} - f_{N_s}(x_p, y_p)|\} \rfloor$$

    IF ( $\varepsilon_V^\infty$ <= precision) THEN DO

        Replace the present model associated to $V$

        by this upper order model

        order $\leftarrow$ order + 1

    END

END

    END

    RETURN ( $\hat{z}$ )

END

In this algorithm version, the maximum aggregating order for models is bounded, but it can be envisioned not to restrict it so and to keep only analytic coefficients up to the order 3, knowing that the upper order coefficients give a less important contribution to the modeling if it is referred to the formulation of Taylor residuals.

Before model aggregation, it is not checked if there is one child model can also apply over the data of the neighbor node, because it is probably the case and the other model should also apply on the neighbor data. In that case the aggregation would provide the equivalent of a simple averaging of the two models even if higher order coefficients are excluded.

At the end in order to validate the approach, it should be assessed in addition that if the barycentric averaging can give a similar result to a least squares linear estimation on data linearized at the corresponding order to the one provided by the aggregation used in the previous algorithm.

If it is referred to the representation in the wavelet form used for illustrating what kind of results could be awaited in applying this approach in image modeling, it should be possible to simplify the algorithm that have just been presented, not in trying to make an image piecewise linear decomposition but by starting aggregation over the piecewise constant representation delivered by the image multi-resolution averaging, which is a step function more or less at a mean quadratic error.

More generally, it can be noted that the progressive decomposition of a global linear model into piecewise linear models stops when the noise influence has been monitored by the least squares method over all the pieces and that the distinction between irregularities and singularities is made by aggregating models at upper interpolation order up to that the approximation error no more fulfills the expected precision : in this case, there is locally a singularity that the method does not succeed to handle.

# 3. Calculus of descriptors

## 3.1. Describing visual shapes

At the end of the segmentation, an image appears in the form of a visual element decomposition that has the following properties :

- each element can be represented as a functional or a collection of functionals leaning on a connected subset of the image domain ;

- each of these functionals can be approached by a finite order polynomial more or less an error ;

- the set of the functional domains makes a partition of the image bi-dimensional domain.

It can then be distinguished two kinds of shape descriptors:

- those that are associated to the piecewise image partition;

- those that are proper to the functionals that are leaning on these pieces that can be approached by a finite order polynomial.

The functional and its domain constitute a simple visual element that we will again name viseme.

## 3.2. Descriptors associated to the shape domains

### 3.2.1. Generalized moments of a shape domain

The moments of the domain $V$ at the order 0, 1, 2 and 3 are given in a continuous way by the following formulas:

- $M(0) = \iint_V dy dx$ ,

area of the domain;

- $M(x) = \iint_V x\, dy dx$ ,     $M(y) = \iint_V y\, dy dx$ ,

from which the gravity center can be extracted;

- $M(x^2) = \iint_V x^2 \, dy\, dx$ , $M(xy) = \iint_V xy \, dy\, dx$ , $M(y^2) = \iint_V y^2 \, dy\, dx$ ,

describe the domain inertia ellipsoid in the observation coordinate system;

- $M(x^3) = \iint_V x^3 \, dy\, dx$ , $\quad M(x^2 y) = \iint_V x^2 y \, dy\, dx$ ,
- $M(xy^2) = \iint_V xy^2 \, dy\, dx$ , $\quad M(y^3) = \iint_V y^3 \, dy\, dx$ ,

expressing the various domain asymmetries according to the axes of the coordinate system.

From the order 2 moments are deduced the main inertia axis and its angle relatively to the observation coordinate system. This one is computed more or less 180° and the uncertainty around 360° is solved by examining the sign o the asymmetry according to the main axis after having moved in the own coordinate system of the domain.

### 3.2.2. Moments calculus over a cellular representation

The discretized expression of generalized moments on a regularly sampled mesh is the following one:

- $M(0) = \sum_{x,y \in V} dm$ ;
- $M(x) = \sum_{x,y \in V} x \, dm$ and $M(y) = \sum_{x,y \in V} y \, dm$ ;
- $M(x^2) = \sum_{x,y \in V} x^2 \, dm$ , $M(xy) = \sum_{x,y \in V} xy \, dm$ , $M(y^2) = \sum_{x,y \in V} y^2 \, dm$ ;
- $M(x^3) = \sum_{x,y \in V} x^3 \, dm$ , $M(x^2 y) = \sum_{x,y \in V} x^2 y \, dm$ , $M(xy^2) = \sum_{x,y \in V} xy^2 \, dm$
  , $M(y^3) = \sum_{x,y \in V} y^3 \, dm$ ,
- where $dm = 1$ .

### 3.2.3. Translation of moments to the domain gravity center

$M(0)$ represents the surface area $S$ of the domain.

The gravity center coordinates are computed with the help of order 1 moments:

- the abscissa of the gravity center $x_G = M(x)/S$ ;
- the ordinate of the gravity center $y_G = M(y)/S$ .

The translation invariance of moments can be deduced by their correction according to the domain gravity center coordinates.

The values of higher order moments in the new coordinate system become:

- $M(X^2) = M(x^2) - x_G^2 S$
- $M(XY) = M(xy) - x_G y_G S$
- $M(Y^2) = M(y^2) - y_G^2 S$
- $M(X^3) = M(x^3) - 3 x_G^2 M(X^2) - y_G^3 S$
- $M(X^2 Y) = M(x^2 y) - y_G M(X^2) - 2 x_G M(XY) - x_G^2 y_G S$
- $M(XY^2) = M(xy^2) - x_G M(Y^2) - 2 y_G M(XY) - x_G y_G^2 S$
- $M(Y^3) = M(y^3) - 3 x_G^2 M(Y^2) - y_G^3 S$

### 3.2.4. Rotation of moments in the domain own coordinate system

From the order 2 moments can be deduced the inertia axes $u_1, u_2$ of the domain:

$$M(u_1^2) = \frac{1}{2}\left(M(X^2) + M(Y^2) + \sqrt{(M(X^2) - M(Y^2))^2 + 4 M(XY)^2}\right)$$

$$M(u_2^2) = \frac{1}{2}\left(M(X^2) + M(Y^2) - \sqrt{(M(X^2) - M(Y^2))^2 + 4 M(XY)^2}\right)$$

This operation returns to assimilate the domain to its inertia ellipsoid of major axis $\vec{u_1}$ and minor axis $\vec{u_2}$ .

The rotation angle that enables to move into its domain own coordinate system $(\overrightarrow{X_G u_1}, \overrightarrow{Y_G u_2})$ , is equal to:

$$\theta(\vec{X}, \vec{u_1}) = arctan\left(\frac{M(u_1^2) - M(X^2)}{M(XY)}\right) \text{ more or less } \pi.$$

Actually, it can be deduced from the domain inertia ellipsoid for the major axis only a direction but not an orientation.

The crossed moment $M(u_1 u_2)$ gets null and the order 3 moments in the directions $u_1$ and $u_2$ become:

$$M(u_1^3) = \cos^3\theta \cdot M(X^3) + 3 \cdot \sin\theta \cdot \cos^2\theta \cdot M(X^2 Y)$$
$$+ 3 \cdot \sin^2\theta \cdot \cos\theta \cdot M(XY^2) + \sin^3\theta \cdot M(Y^3)$$
$$M(u_1^2 u_2) = -\sin\theta \cdot \cos^2\theta \cdot M(X^3) + \cos^3\theta \cdot M(X^2 Y)$$
$$- 2 \cdot \sin^2\theta \cdot \cos\theta \cdot M(X^2 Y) + 2 \cdot \sin\theta \cdot \cos^2\theta \cdot M(XY^2)$$
$$- \sin^3\theta \cdot M(XY^2) + \sin^2\theta \cdot \cos\theta \cdot M(Y^3)$$
$$M(u_1 u_2^2) = \sin^2\theta \cdot \cos\theta \cdot M(X^3) - 2 \cdot \sin\theta \cdot \cos^2\theta \cdot M(X^2 Y)$$
$$+ \sin^3\theta \cdot M(X^2 Y) + \cos^3\theta \cdot M(XY^2)$$
$$- 2 \cdot \sin\theta \cdot \cos^2\theta \cdot M(XY^2) + \sin\theta \cdot \cos^2\theta \cdot M(Y^3)$$
$$M(u_2^3) = -\sin^3\theta \cdot M(X^3) + 3 \cdot \sin^2\theta \cdot \cos\theta \cdot M(X^2 Y)$$
$$- 3 \cdot \sin\theta \cdot \cos^2\theta \cdot M(XY^2) + \cos^3\theta \cdot M(Y^3)$$

The orientation of axes $\vec{u_1}$ et $\vec{u_2}$ is given by making positive the moment $M(u_1^3)$.

It is equal to give to $\vec{u_1}$ the orientation that provides the highest asymmetry according to this axis.

The domain main axis asymmetry is a characteristics that belongs to it and which is stable once determined the orientations of inertia axes.

The uncertainty more or less 360° is then solved in the following way:

$$M(u_1^3) < 0 \Rightarrow \theta = \theta + \pi$$

and the order 3 moments are updated by a single sign change:

$$M(u_1^3) = -M(u_1^3)$$
$$M(u_1^2 u_2) = -M(u_1^2 u_2)$$
$$M(u_1 u_2^2) = -M(u_1 u_2^2)$$
$$M(u_2^3) = -M(u_2^3)$$

These values are representing the various domain asymmetries in its own coordinate system.

### 3.2.5. Invariant characteristics of the surface pieces domains

It has been just described how to get :

- the gravity center coordinates $x_G, y_G$ of the shape domain;
- the angle $\theta$ of the shape domain in the image coordinate system.

The other moments calculated in the shape own coordinate system $(x_G, y_G, \theta)$ enable to have at one's disposal spatial characteristics about this domain that are translation and rotation invariant in the shooting plane.

These characteristics are:

- $M(0)$ the domain area ;
- $M(u_1^2), M(u_2^2)$ the domain inertia values;
- $M(u_1^3), M(u_1^2 u_2), M(u_1 u_2^2), M(u_2^3)$ the domain asymmetries.

These characteristics become homothety invariant by dropping the domain area and by normalizing the remaining moments by the major inertia axis value. It will remain:

- $\epsilon = \dfrac{M(u_2^2)}{M(u_1^2)}$ the shape domain eccentricity;
- $\dfrac{M(u_1^3)}{M(u_1^2)}, \dfrac{M(u_1^2 u_2)}{M(u_1^2)}, \dfrac{M(u_1 u_2^2)}{M(u_1^2)}, \dfrac{M(u_2^3)}{M(u_1^2)}$ the scaled asymmetries.

It should be added to the shape localization information the scale factor constituted by the major inertia axis $M(u_1^2)$ in order to correctly replace this one in the image plane.

## *3.3. Descriptors of shape rendering*

### 3.3.1. Analytic expansion of pieces

In the aim of providing measures enabling to describe the surface pieces coming from the piecewise regular decomposition of a digitized image, it will be focused on models of at most order 3 for computing these measures or their expansion up to order 3.

For each subset $V_j$, $j \in 1,n(3)$ of the image partition got by regular decomposition up to order 3, each spectral band can be approximated by the cubic:

$$\hat{z}_l^j = a_{1,l}^j + a_{x,l}^j \cdot x + a_{y,l}^j \cdot y + a_{x^2,l}^j \cdot x^2 + a_{xy,l}^j \cdot xy + a_{y^2,l}^j \cdot y^2 + a_{x^3,l}^j \cdot x^3 + a_{x^2y,l}^j \cdot x^2 y + a_{xy^2,l}^j \cdot xy^2 + a_{y^3,l}^j \cdot y^3$$

By forgetting the index numbers $j$ and $l$, the expression makes simpler as:

$$\hat{z} = a_1 + a_x \cdot x + a_y \cdot y + a_{x^2} \cdot x^2 + a_{xy} \cdot xy + a_{y^2} \cdot y^2 + a_{x^3} \cdot x^3 + a_{x^2 y} \cdot x^2 y + a_{xy^2} \cdot xy^2 + a_{y^3} \cdot y^3$$

### 3.3.2. Translation of the analytic expressions towards the piece gravity centers

If $\begin{pmatrix} \bar{x}_j \\ \bar{y}_j \end{pmatrix}$ is the gravity center of the piece $V_j$, its estimate in the $l^{nth}$ spectral band will be equal to

$\hat{z}_l^j$ = order 3 estimates at $\begin{pmatrix} \bar{x}_j \\ \bar{y}_j \end{pmatrix}$ that will be written in a simplified manner $\begin{pmatrix} \bar{z} \\ \bar{x} \\ \bar{y} \end{pmatrix}$. That is to say that our focus is moving from the digitized image to its model at the neighborhood of the domain gravity center of each image piece.

For doing it, place ourselves in the coordinate system of the tangent plane of the spectral band estimate centered at the gravity center $M_j$ of the piece $V_j$.

### 3.3.3. Description of a shape in the coordinate system of its tangent plane

In order to achieve it, it must be performed a translation to $\begin{pmatrix} \bar{z} \\ \bar{x} \\ \bar{y} \end{pmatrix}$, then rotation defined from the equation of the directrix plane deduced from the order 1 coefficients of the estimate, that is such as the directrix plane gets for equation at $\begin{pmatrix} \bar{z} \\ \bar{x} \\ \bar{y} \end{pmatrix}$ $Z = a_x \cdot X + a_y \cdot Y$ where $Z = z - \hat{z}$, $X = x - \bar{x}$, $Y = y - \bar{y}$ and therefore for directrix coefficients:

- $\tan(\theta_{XZ}) = a_x$ in the plane $(\begin{pmatrix} \bar{z} \\ \bar{x} \\ \bar{y} \end{pmatrix}, \vec{X}, \vec{Z})$ ;

- $\tan(\theta_{YZ}) = a_y$ in the plane $(\begin{pmatrix} \bar{z} \\ \bar{x} \\ \bar{y} \end{pmatrix}, \vec{Y}, \vec{Z})$ .

So moving to the new coordinate system associated to the tangent plane of the surface centered at the domain gravity center of the piece $V_j$ returns in applying the following linear transform on every points of the concerned surface :

$$\begin{pmatrix} Z \\ X \\ Y \end{pmatrix} = \begin{pmatrix} \cos(\theta_{XZ}) & \sin(\theta_{XZ}) & 0 \\ -\sin(\theta_{XZ}) & \cos(\theta_{XZ}) & 0 \\ 0 & 0 & 1 \end{pmatrix} \begin{pmatrix} \cos(\theta_{YZ}) & 0 & \sin(\theta_{YZ}) \\ 0 & 1 & 0 \\ -\sin(\theta_{YZ}) & 0 & \cos(\theta_{YZ}) \end{pmatrix} \begin{pmatrix} z - \bar{z} \\ x - \bar{x} \\ y - \bar{y} \end{pmatrix},$$

that is also:

$$\begin{pmatrix} Z \\ X \\ Y \end{pmatrix} = \begin{pmatrix} \cos(\theta_{XZ})\cos(\theta_{YZ}) & \sin(\theta_{XZ}) & \cos(\theta_{XZ})\sin(\theta_{YZ}) \\ -\sin(\theta_{XZ})\cos(\theta_{YZ}) & \cos(\theta_{XZ}) & -\sin(\theta_{XZ})\sin(\theta_{YZ}) \\ -\sin(\theta_{YZ}) & 0 & \cos(\theta_{YZ}) \end{pmatrix} \begin{pmatrix} z - \bar{z} \\ x - \bar{x} \\ y - \bar{y} \end{pmatrix}.$$

As $\tan(\theta) = \dfrac{\sin(\theta)}{\cos(\theta)}$ and $\cos(\theta) = \sqrt{\dfrac{1}{1 + \tan(\theta)^2}}$, it would be advised to use the formula $\sin(\theta) = \tan(\theta) \cdot \cos(\theta)$ in order to keep unchanged the tangent sign when the rotation matrix will be built.

In the coordinate system linked to the tangent plane centered at the piece gravity center, the surface equation of the local image estimate takes the form of the reduced expression :

$$\hat{Z} = a_{X^2} \cdot X^2 + a_{XY} \cdot XY + a_{Y^2} \cdot Y^2 + a_{X^3} \cdot X^3 + a_{X^2Y} \cdot X^2 Y + a_{XY^2} \cdot XY^2 + a_{Y^3} \cdot Y^3$$

Because it corresponds also to the finite expansion of the surface equation at the point $M$ that can be written by using the Taylor formula up to the order 3 :

$$\hat{z} = \bar{z} + \frac{\delta z}{\delta x}(\bar{x}, \bar{y}) \cdot x + \frac{\delta z}{\delta y}(\bar{x}, \bar{y}) \cdot y + \frac{\delta^2 z}{\delta^2 x}(\bar{x}, \bar{y}) \cdot x^2 + \frac{\delta^2 z}{\delta x \delta y}(\bar{x}, \bar{y}) \cdot xy + \frac{\delta^2 z}{\delta^2 y}(\bar{x}, \bar{y}) \cdot y^2$$
$$+ \frac{\delta^3 z}{\delta^3 x}(\bar{x}, \bar{y}) \cdot x^3 + \frac{\delta^3 z}{\delta^2 x \delta y}(\bar{x}, \bar{y}) \cdot x^2 y + \frac{\delta^3 z}{\delta x \delta^2 y}(\bar{x}, \bar{y}) \cdot xy^2 + \frac{\delta^3 z}{\delta^3 y}(\bar{x}, \bar{y}) \cdot y^3 + o^3(x, y)$$

To express the local image estimate in the coordinate system associated to the tangent plane centered at the piece gravity center turns then in canceling the constant term and the gradient of the function in its polynomial expression.

### 3.3.4. Shape own coordinate system and reduction of its analytic expression

Carry on the reduction of this equation by calculating its expression in the own coordinate system $(M, \vec{u}, \vec{v})$ of the quadric $a_{X^2} \cdot X^2 + a_{XY} \cdot XY + a_{Y^2} \cdot Y^2$ in $(M, \vec{X}, \vec{Y})$.

This quadric expression can be also written :

$$a_{X^2} \cdot X^2 + a_{XY} \cdot XY + a_{Y^2} \cdot Y^2 = (X \ Y) \begin{pmatrix} \frac{\delta^2 Z}{\delta^2 X}(0,0) & \frac{\delta^2 Z}{\delta X \delta Y}(0,0) \\ \frac{\delta^2 Z}{\delta X \delta Y}(0,0) & \frac{\delta^2 Z}{\delta^2 Y}(0,0) \end{pmatrix} \begin{pmatrix} X \\ Y \end{pmatrix} = \begin{pmatrix} X \\ Y \end{pmatrix}^T H_Z \begin{pmatrix} X \\ Y \end{pmatrix} \ ,$$

where $H_Z$ is the hessian matrix of $Z$.

As it is a real symmetric matrix, it can be also written $H_Z = P \begin{pmatrix} \lambda_u & 0 \\ 0 & \lambda_v \end{pmatrix} P^{-1}$, where $\lambda_u$ and $\lambda_v$ are the eigenvalues of the hessian of $Z$ and $P$ their eigenvectors matrix.

These eigenvalues have got the expressions :

$$\lambda_u = \frac{1}{2}(a_{X^2} + a_{Y^2} + \sqrt{(a_{X^2} - a_{Y^2})^2 + 4 a_{XY}^2}) \text{ and } \lambda_v = \frac{1}{2}(a_{X^2} + a_{Y^2} - \sqrt{(a_{X^2} - a_{Y^2})^2 + 4 a_{XY}^2})$$

The matrix $P$ is the rotation matrix of angle : $\theta_{Xu}=\arctan(\frac{\lambda_u-a_{X^2}}{a_{XY}})$ more or less $\pi$ in the tangent plane, that can then be written : $P=\begin{pmatrix} \cos(\theta_{Xu}) & \sin(\theta_{Xu}) \\ -\sin(\theta_{Xu}) & \cos(\theta_{Xu}) \end{pmatrix}$ .

After having performed this rotation in the tangent plane going through $M$, it is obtained a new equation for representing the estimate:

$$\hat{Z}_\theta = \lambda_u \cdot u^2 + \lambda_v \cdot v^2 + a_{u^3} \cdot u^3 + a_{u^2 v} \cdot u^2 v + a_{uv^2} \cdot uv^2 + a_{v^3} \cdot v^3$$, that can be normalized by doing the

scaling : $\hat{w} = u^2 + \frac{\lambda_v}{\lambda_u} \cdot v^2 + \frac{a_{u^3}}{\lambda_u} \cdot u^3 + \frac{a_{u^2 v}}{\lambda_u} \cdot u^2 v + \frac{a_{uv^2}}{\lambda_u} \cdot uv^2 + \frac{a_{v^3}}{\lambda_u} \cdot v^3$ .

If $\lambda_v / \lambda_u > 0$ then the quadric reduced equation is the one of an elliptic paraboloid, else it is the one of a hyperbolic paraboloid. The coefficients $a_{u^3}$, $a_{u^2 v}$, $a_{uv^2}$ and $a_{v^3}$ are measuring the asymmetric distortions of these surfaces around these quadrics.

### 3.3.5. Invariant characteristics of the shapes associated to the surface pieces

The shape rendering characteristics coming from the piecewise regular decomposition are resting on the gravity centers of the shape domains, that are the planar vectors $\begin{pmatrix} x_G \\ y_G \end{pmatrix}$, that will be re-written $\begin{pmatrix} \bar{z} \\ \bar{x} \\ \bar{y} \end{pmatrix}$ by including the mean values of the spectral bands on the domain V of each piece, that is where $\bar{z} = \begin{pmatrix} \bar{z}_1 \\ \bar{z}_2 \\ \vdots \\ \bar{z}_{N_S} \end{pmatrix}$ .

It will get then in position in the collection of the $N_S$ tangent planes at point $\begin{pmatrix}\overline{z}\\\overline{x}\\\overline{y}\end{pmatrix}$ by performing the following rotations of angles:

- $\theta_{XZ}^l = \arctan(a_x^l)$ in the plane $(\begin{pmatrix}\overline{z}^l\\\overline{x}\\\overline{y}\end{pmatrix}, \vec{X}, \vec{Z})$, for $l \in \{1, \cdots, N_S\}$ ;

- $\theta_{YZ}^l = \arctan(a_y^l)$ in the plane $(\begin{pmatrix}\overline{z}^l\\\overline{x}\\\overline{y}\end{pmatrix}, \vec{Y}, \vec{Z})$, for $l \in \{1, \cdots, N_S\}$ .

Then it will be done in each tangent plane the rotation of angle $\theta_{Xu} = \arctan(\dfrac{\lambda_u - a_{X^2}}{a_{XY}})$ in order to arrive in the own coordinate system of the quadric $a_{X^2} \cdot X^2 + a_{XY} \cdot XY + a_{Y^2} \cdot Y^2$.

As for the shape domain, one will have at its disposal after normalization of the characteristics invariant to similarities for the shape rendering in each spectral band:

- $\dfrac{\lambda_v}{\lambda_u}$ in the space $(\begin{pmatrix}\overline{z}\\\overline{x}\\\overline{y}\end{pmatrix}, \vec{X}, \vec{Y}, \vec{Z})$ ;

- $\dfrac{a_{u^3}}{\lambda_u}, \dfrac{a_{u^2v}}{\lambda_u}, \dfrac{a_{uv^2}}{\lambda_u}, \dfrac{a_{v^3}}{\lambda_u}$ in the space $(\begin{pmatrix}\overline{z}\\\overline{x}\\\overline{y}\end{pmatrix}, \vec{X}, \vec{Y}, \vec{Z})$ .

# 4. Building perceptual groupings

## *4.1. Introduction*

At the beginning of XX[th] century, psychologists have built a holistic theory for perception named Gestalttheorie. It was about looking for what are the laws that are ruling human visual perception. Psycho-sensory experiments have allowed to find six main laws:

- the good shape law: a shapeless part set tends towards being first perceived as a simple shape, symmetric and stable;
- the law of continuity: gathered points are perceived the expansion of each others as a single shape;
- the law of proximity: the points the nearest from each others are first and foremost gathered;
- the similarity law: if they are not enough near, the most similar between them are gathered into a single shape;
- the law of common destiny: moving parts having the same trajectory are perceived as belong to the same shape;
- the law of familiarity: the most familiar shapes are perceived as being the most significant ones.

This perception approach relies on the spatial structuring of visual shapes and enables to envision and develop partially hidden pattern recognition methods by identifying wholly or partly an object in one's view field.

The computation of digital descriptors over visual shapes (good shape law) provided by an image segmentation (law of continuity) allows to recognize only fully visible objects in the view field. Which is hardly possible knowing that human vision only provides a planar view on a tridimensional physical universe by using a projective transformation.

In order to solve this issue, it is proposed to adapt the attribute calculus so as to include some laws of shape perception by aggregating the closest shape attributes (law of proximity) along the treelike path of the shapes that are belonging to an image.

This approach is holding simultaneously the advantages of the statistical pattern recognition (similarity law) as well as those of structural pattern recognition (law of familiarity). It has been already tested in the framework of a past application for facial recognition in planar color images.

The monitoring of moving groupings image by image in a video should illustrate he law of common destiny.

The attribute calculus implies that the object of interest does not include some hidden part and may appear in the form of a single connected component. These conditions are generally not satisfied: only one part of a tridimensional object can be viewed by a single observer and image segmentation may provide a decomposition of the observed object into multiple connected components. In order to solve this issue, it is proposed to insert an intermediate computing step based on the following holistic scheme: neighbor regions are successively aggregated in the form of compound regions until being able to rebuild as wholly as possible the interesting object. The calculus of descriptors is so generalized to compound objects and an object will then appear as a series of components nested the ones into the others, that is to say a series of complex shapes. Each one among them will have at one's disposal its own attribute vector and should be individually identified: so it will be enabled to recognize an object without seeing it entirely by identifying part of the parts that are composing it.

## *4.2. Shape aggregation*

Moment-based attribute calculus assumes that objects to recognize are viewed as a whole. It is not actually fulfilled with shape recognition applications. Statistical learning made from the recording of several views of a same object observed with distinct positions and attitudes, allows to create a sufficient dense set of feature data for expecting to recognize any incoming new view standing near some previously recorded ones.

But the moment-based attribute calculus using low-order formulas allows only to identify compact shapes. On another side, the image piecewise regular segmentation would provide a richer object decomposition in more numerous regions. To take benefit of this property, it is proposed to reuse the shape aggregative process used by the past in facial recognition ([30]). Usually region merging proposes to aggregate adjacent regions. To avoid building up a regional adjacency graph, an alternate aggregative approach based on region nearness has been developed. During region aggregation, attribute calculus is performed without centering, normalizing and scaling. As moments are integral formulas, attributes can be then straightforwardly summed up before being post-processed. Using gravity center information, a regional gravity center quad-tree can easily be built and allows to merge nearest regions using a bottom-up tree traversal. As the merging process is recursively repeated up to tree root, the various objects of interest belonging to the image can then be successively decomposed in a nested set of regions.

After attribute summation, the features are centered, normalized and scaled at every decomposition level of an image model. So an image object can be then described as a variable length series of feature vectors. At training time, the labels are applied on vector series allowing to provide a recognition step running on wholly or partially visible objects.

## *4.3. Facial recognition example*

The statistical learning step made from the recording of several views of a same person, observed according to different positions and attitudes, allows to create a dense set of characteristics data in the aim to recognize a person viewed in a situation close to those that have been already registered.

The example here presented is based on color image segmentation. The shape aggregation method takes in account most of the region groupings that can appear in the decomposition of a face. As the aggregation process is recursively repeated up to the top of the tree of the region gravity centers, the face of a person can be progressively recomposed as a set of nested regions.

In this example, only shape domain descriptors are taken in account. After having summed up the attributes of the whole regions, the characteristics can be centered, normalized and scaled at each decomposition level of a face model. So a face can be described by a variable length series of characteristics vectors. During learning, labels are assigned to these vector series in order to set up a facial recognition stage.

An example of shape progressive aggregation is shown underneath. The attribute list of the various objects provided by aggregation of such images is next after displayed. They are consisting in localization and shape attributes. The gravity center coordinates, the angle and the scale factor are localization attributes of the corresponding object in the shooting coordinate system. The surface, the eccentricity and the asymmetries compose the shape attributes of the object. They are centered, normalized and scaled values, which are independent from any translation, rotation and homothety in the capturing coordinate system.

In this specific application, the decision of authenticating or identifying one person among several is taken according to the label gathering the maximum number of recognized simple or compound shapes.

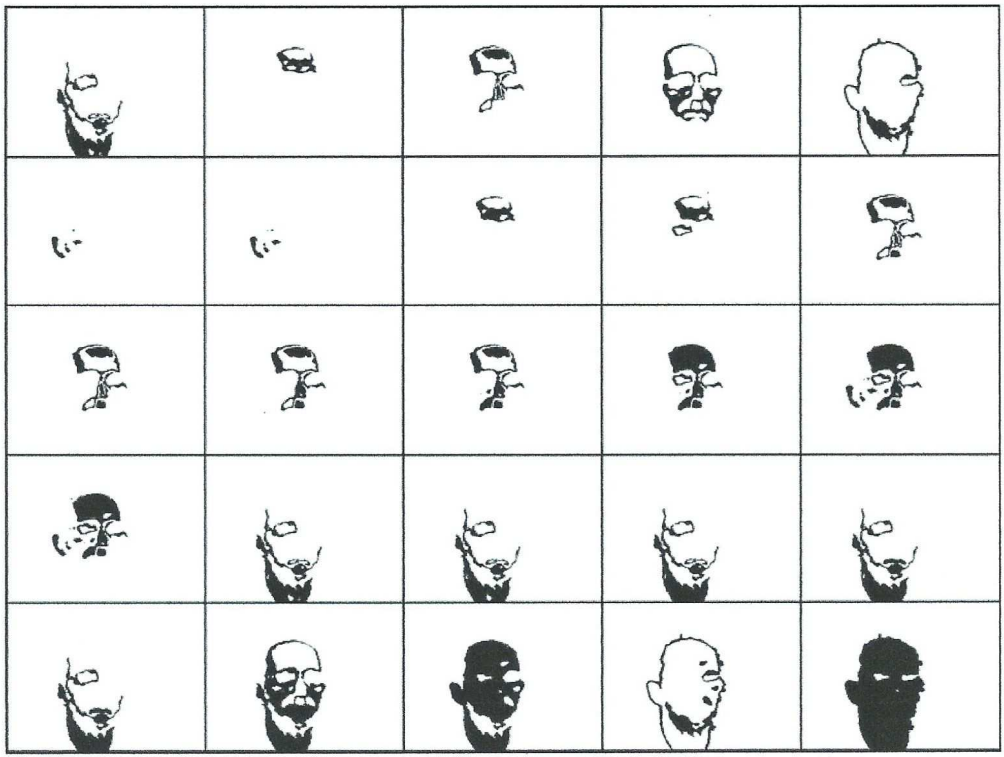

**Figure 4 : Hierarchical aggregation of regions**

| Object n° | Center Abs. | Center Ord | Angle | Scale | Surface | Eccentric. | 1st Asym. | 2nd Asym. |
|---|---|---|---|---|---|---|---|---|
| 0 | 169.279 | 190.920 | 273.545 | 59.559 | 0.4747 | 0.7057 | 0.8579 | 0.3849 |
| 1 | 168.971 | 160.698 | 238.319 | 39.068 | 0.7220 | 0.7628 | 0.7980 | 0.5896 |
| 2 | 176.130 | 121.729 | 77.629 | 34.416 | 0.6487 | 0.7082 | 0.7952 | 0.1673 |
| 3 | 178.040 | 105.748 | 15.728 | 20.979 | 0.7661 | 0.5030 | 0.1476 | 0.3890 |
| 4 | 153.887 | 226.375 | 249.587 | 47.281 | 0.6608 | 0.5542 | 0.8516 | 0.4414 |
| 5 | 109.085 | 173.296 | 350.625 | 16.242 | 0.6306 | 0.7493 | 1.0041 | 0.2672 |
| 6 | 109.190 | 173.068 | 348.652 | 16.265 | 0.6320 | 0.7594 | 1.0039 | 0.3471 |
| 7 | 177.521 | 106.380 | 14.977 | 20.360 | 0.8164 | 0.5211 | 0.3752 | 0.4272 |
| 8 | 173.352 | 112.506 | 339.856 | 20.794 | 0.8888 | 0.8004 | 0.4257 | 0.5166 |
| 9 | 176.674 | 126.672 | 79.848 | 37.619 | 0.6162 | 0.6263 | 0.7472 | 0.2434 |
| 10 | 176.769 | 127.016 | 79.740 | 37.723 | 0.6167 | 0.6225 | 0.7359 | 0.2379 |
| 11 | 177.078 | 128.370 | 79.654 | 37.591 | 0.6319 | 0.6125 | 0.6774 | 0.2259 |
| 12 | 176.152 | 132.086 | 84.307 | 38.740 | 0.6347 | 0.5916 | 0.5893 | 0.2757 |
| 13 | 175.141 | 125.018 | 85.145 | 33.952 | 0.9060 | 0.6508 | 0.8530 | 0.3482 |
| 14 | 168.515 | 129.846 | 115.188 | 37.001 | 0.8765 | 0.7405 | 0.8251 | 0.2190 |
| 15 | 168.470 | 129.799 | 115.066 | 36.988 | 0.8773 | 0.7421 | 0.8255 | 0.2340 |
| 16 | 153.939 | 226.471 | 249.548 | 47.319 | 0.6609 | 0.5533 | 0.8515 | 0.4404 |
| 17 | 153.982 | 226.550 | 249.521 | 47.341 | 0.6610 | 0.5527 | 0.8518 | 0.4396 |
| 18 | 154.134 | 227.217 | 249.691 | 47.404 | 0.6645 | 0.5487 | 0.8589 | 0.4397 |
| 19 | 154.838 | 226.784 | 249.857 | 46.836 | 0.6810 | 0.5600 | 0.8577 | 0.4307 |
| 20 | 154.742 | 226.803 | 249.846 | 46.771 | 0.6829 | 0.5613 | 0.8577 | 0.4324 |
| 21 | 160.977 | 197.839 | 264.764 | 53.025 | 0.8036 | 0.5928 | 0.5511 | 0.4162 |
| 22 | 163.727 | 172.865 | 92.444 | 57.546 | 0.9308 | 0.5386 | 0.6141 | 0.2124 |
| 23 | 172.523 | 188.593 | 275.887 | 58.201 | 0.5090 | 0.7112 | 0.8155 | 0.5023 |
| 24 | 165.788 | 176.550 | 92.379 | 58.052 | 1.0544 | 0.5853 | 0.3648 | 0.2288 |

**Tableau 1 : Attribute list of aggregated regions**

## 4.4. Shape aggregation algorithms

### 4.4.1. Aggregation of shape rendering descriptors

In the concerned case, it is unnecessary to build the gravity center tree of all the regions that make up the image, because it is the same that is provided by the image piecewise regular decomposition.

It will be just enough to modify the piecewise regular decomposition algorithm used for aggregating shape descriptors by deleting the test about the order not to overcome concerning them and to restrict the algorithm in only keeping the order 3 coefficients generated by barycentric combination. Here is the piecewise regular decomposition algorithm revised in this intent while introducing the explicit handling of the data treelike structure:

BEGIN

    order ← 1

    tree ← WHITE

    CALL Piecewise regular decomposition ( $V$ , tree, order, precision)

END

PROCEDURE Piecewise regular decomposition ( $V$ , root, order, precision)

BEGIN

    Compute the linear estimate of $V$ according to least squares method:

$$\hat{z} = \begin{pmatrix} f_1 \\ \vdots \\ f_l \\ \vdots \\ f_{N_s} \end{pmatrix} \begin{pmatrix} 1 \\ x \\ y \end{pmatrix} \text{ where the } f_l \text{ are row vectors resulting from the linear estimation}$$

    Compute the maximum approximation error:

$$\varepsilon_V^\infty = \lfloor Max_{p \in V} \{|z_{1,p} - f_1(x_p, y_p)|, \cdots, |z_{N_s, p} - f_{N_s}(x_p, y_p)|\} \rfloor$$

    IF ( $\varepsilon_V^\infty$ > precision) THEN DO

Compute the error histogram

IF (mono-modal histogram)

THEN $V_S \leftarrow \{\sqrt{Card(V)}$ points of strongest error $\}$

ELSE DO

    Define as threshold the histogram highest valley

    $V_S \leftarrow \{p \in V / \epsilon_p^\infty > threshold\}$

END

Compute the coordinates $\bar{x}$ and $\bar{y}$ of the gravity center

and the variance $\sigma_{x^2}$ and covariance $\sigma_{xy}$ of $V_S$

Divide the set $V$ into two subsets:

$V_- = \{p \in V / (x-\bar{x}) \cdot \sigma_{xy} - (y-\bar{y}) \cdot \sigma_{x^2} < 0\}$

$V_+ = \{p \in V / (x-\bar{x}) \cdot \sigma_{xy} - (y-\bar{y}) \cdot \sigma_{x^2} \geq 0\}$

IF (TERMINAL(root) THEN FISSION(root)

$\widehat{z_-} \leftarrow$ Piecewise regular decomposition ( $V_-$ , CHILD(root, LEFT), order, precision)

$\widehat{z_+} \leftarrow$ Piecewise regular decomposition ( $V_+$ , CHILD(root, RIGHT), order, precision)

$\widehat{z_-} \leftarrow$ VALUE(CHILD(root, LEFT))

$\widehat{z_+} \leftarrow$ VALUE(CHILD(root, RIGHT))

Compute the barycentric combination:

$$\hat{z}=\frac{(x-\overline{x_-})\cdot(\overline{x_+}-\overline{x_-})+(y-\overline{y_-})\cdot(\overline{y_+}-\overline{y_-})}{(\overline{x_+}-\overline{x_-})^2+(\overline{y_+}-\overline{y_-})^2}\cdot\widehat{z_-}+\frac{(\overline{x_+}-x)\cdot(\overline{x_+}-\overline{x_-})+(\overline{y_+}-y)\cdot(\overline{y_+}-\overline{y_-})}{(\overline{x_+}-\overline{x_-})^2+(\overline{y_+}-\overline{y_-})^2}\cdot\widehat{z_+}$$

    Compute the approximation error:

$$\varepsilon_V^\infty=\lfloor Max_{p\in V}\{|z_{1,p}-f_1(x_p,y_p)|,\cdots,|z_{N_s,p}-f_{N_s}(x_p,y_p)|\}\rfloor$$

    VALUE(root) ← $\hat{z}$    truncated at the order 3

    order ← order + 1

  END

  RETURN

END

### 4.4.2. Aggregation of shape domain descriptors

At the difference of the previous computation of shape domain descriptors, this computing step must not being stopped at the level of terminal nodes, but must be carried on up to the decomposition tree root. It will be only after having reached it that the shape domain descriptors could be entered and normalized up to the tree root. It is the same with the shape rendering descriptors where the barycentric combination truncated at order 3 will be used for computing the coefficients of the aggregated up to the third expansion order.

## *4.5. Similarity in multidimensional hierarchical modeling*

For implementing a shape recognition process based on visual descriptors, it is proposed to focus our attention on multidimensional hierarchical modeling techniques.

The modeling of multidimensional data sets by the means of hierarchical treelike structures implies that the sets are managed using an ultra-metric distance. Concerning regularly divided spaces, the structuring distance is the Hausdorff distance: the distance between two points in the space is the size of the smallest subset in the space that is holding together these two points. It is not a distance in the conventional meaning that allows to measure the distance between two points from the space but a pseudo-distance used for comparing subsets between them.

As the characteristics describing objects are normalized, it might be possible to use an unweighted euclidean distance for defining a similarity measure. But it would be rather considered each characteristic vector as a point belonging to a multidimensional space of dimension equal to the number of characteristics computed over the image objects. Consequently the characteristics coordinates are straightly managed by the treelike structure representing the learning set and this enables to compare directly each new vector with anyone vector registered during learning. So conditions are more favorable for assessing similarity between known and unknown objects. This skill will be especially useful for detecting and identifying moving objects in image series and eventually for enriching a learning base with the characteristics computed on these new objects. Moreover, this similarity measure induced by a treelike structure enables to sort all the objects belonging to a data base according their similarity with a sample vector (query by example) or according their self-similarity (data base sort).

# 5. Building visual dictionaries

## *5.1. Vector quantization of shape descriptors*

For numerically describing simple and compound shapes provided by an image piecewise regular segmentation, we have at our disposal for each shape:

- an attribute vector associated to its shape domain, based on the computation of generalized moments, and consisting in values invariant to the geometrical transformations that can apply on an object in the image plane;
- and for each spectral component of the image, a complementary vector describing the shape rendering as well as invariant but in the functional space of the image.

In order to get a description up to a given precision for this numerical information, it is proposed to use a treelike structure for handling these numerical data vectors ([13],[28]), enabling to both represent images and visual objects attributes so as to perform shape recognition processes using statistical learning.

As a reminder, it is an indexing scheme based on binary tree structures enabling to represent spaces of any dimension regularly divided and used for registering numerical data coming from a learning set that has been labeled or not with the names of objects to be recognized ([11],[12]). For representing a set of numerical data belonging to a $k$ dimension space, the data must be registered into a binary tree where the attribute vectors are used as geographical coordinate keys for retrieving the information from the tree. For localizing a data vector, it must be followed a traversal similar to the one made for searching in a quaternary or octernary tree:

- according to the divide and conquer paradigm, the multidimensional space is divided half by half successively along each space direction until reaching the modeling precision given as a search parameter;
- at each dividing step of the space, the vector coordinates are assessed and the decision to go towards such or such left or right half-space is taken until reaching the required analysis precision in the tree.

Let us take the example of the addition of a normalized vector, as it is the case of shape descriptors, to a binary tree modeling the unitary hyper-cube of dimension $k$ at the precision $r$:

BEGIN

    Initialization of the addition of a normalized vector to a binary tree representing a $2^k$-tree:

vector ← vector to be added to the data already belonging to a given tree

root ← root of the tree in which must be performed the addition

minroot(1 : $k$ ) ← $\{0., 0., \cdots, 0.\}$

maxroot(1 : $k$ ) ← $\{1., 1., \cdots, 1.\}$

level ← level reached in the tree

dimension ← $k$

depth ← $k \cdot r$

CALL Addition of a normalized vector to a binary tree (vector, root, minroot, maxroot,
   0, dimension, depth)

END

PROCEDURE Addition of a normalized vector to a binary tree (vector, root, minroot, maxroot, level, dimension, depth)

BEGIN

   IF ((level <> depth) AND ( NOT BLACK(root))  THEN DO

      Descent in the tree driven by the vector coordinates:

      nudim ← (level% dimension) + 1

      center ← (minroot(nudim) + maxroot(nudim) / 2.

      IF (vector(nudim) < center) THEN maxroot(nudim) ← center, side ← LEFT

      ELSE minroot(nudim) ← center, side ← RIGHT

      IF (TERMINAL(root)) THEN FISSION(root)

CALL Addition of a normalized vector to a binary tree (vector, CHILD(root, side),

minroot, maxroot, level + 1, dimension, depth)

END

ELSE BLACKEN(root)

IF (NOT TERMINAL(root)) THEN MERGE(root)

END

This recursive procedure enables to record the presence of a vector of any dimension in the unitary space of the same dimension by handling this space in the form of a binary tree, that is equal to map

$$\left\{0, \frac{1}{2^r}, \ldots, \frac{2^r-1}{2^r}\right\}^k \text{ into } \left\{0, \frac{1}{2^{kr}}, \ldots, \frac{2^{kr}-1}{2^{kr}}\right\}.$$

This one is created empty using the instruction root ← WHITE, that generates a white tree-node representing an empty space of any dimension at any precision. The tree branches grow progressively at each vector addition with the help of the operator FISSION and two iso-colored children nodes representing two neighbor vectors at precision $r$ are merged into a single node at father level with the help of the operator MERGE. The nodes BLACK and WHITE are tree terminal nodes that tell if some data is present or not in the branch that have been traversed.

Mapped into a binary in such a way, a multidimensional space can be seen as a series of quadrants, octants or hexadecants successively nested the ones into the other ones according to the space dimension. Their union gives back the whole initial space in which they are mapped and a simple binary searching enables to retrieve any point or subset belonging to this space modeled in such a way. The precision is a parameter used during the building or the search of information in this space and corresponds to the traversal depth of the representation tree.

Then a learning data set can be represented as a collection of labeled trees built at a given precision where the labels are the identifiers of the objects to be recognized, indexed by their descriptor vectors computed during image analysis, at the rate of one tree specifically for the domain attributes and as many trees as they are spectral components. More the precision is low, more the quantization will be rough and more it is high, more it will be detailed.

## 5.2. Visual alphabet of simple shapes

The simple shapes provided by an image piecewise regular segmentation consist in:

- an attribute vector describing the domain of a shape, based on the computation of generalized moments;
- and for each image spectral component, a complementary vector describing the rendering of a shape.

The attributes describing the domain of each shape are the values:

- $\epsilon = \dfrac{M(u_2^2)}{M(u_1^2)}$ the eccentricity of the shape domain;

- $\dfrac{M(u_1^3)}{M(u_1^2)}, \dfrac{M(u_1^2 u_2)}{M(u_1^2)}, \dfrac{M(u_1 u_2^2)}{M(u_1^2)}, \dfrac{M(u_2^3)}{M(u_1^2)}$ the normalized asymmetries.

Concerning each image spectral component, the rendering of a shape is described by the series of values: $\dfrac{\lambda_v}{\lambda_u}$, $\dfrac{a_{u^3}}{\lambda_u}, \dfrac{a_{u^2 v}}{\lambda_u}, \dfrac{a_{uv^2}}{\lambda_u}, \dfrac{a_{v^3}}{\lambda_u}$.

From a practical viewpoint, it may restrict a shape to the convex hull of its domain, that is standing in the coefficients $\dfrac{M(u_2^2)}{M(u_1^2)}, \dfrac{M(u_1^3)}{M(u_1^2)}, \dfrac{M(u_2^3)}{M(u_1^2)}$ and in a unique coefficient per spectral component $\dfrac{\lambda_v}{\lambda_u}$ enough for describing a quadric surface.

All these values are independent from the linear transformations that may apply on shapes in the visual space in which they have been captured. After quantization, they can be reduced to finite subsets of normalized values that can be seen as visual alphabets enabling to encode images in the form of series of ideographic characters as it was the case with the first writing systems developed by humanity. According to the precision used for quantifying simple shapes, the alphabet so produced can be more or less rich.

Let they be the descriptors of shape domains as well as their rendering ones, the quantization is performed by building their trees in their own coordinate systems at a precision wished for recognizing them, then building them back during their synthesis.

As the space is unitary, the algorithm presented in the previous paragraph can apply not only for encoding the values of these descriptors, but also the points sampling the shape of each domain so as to be reused during image synthesis. So each shape domain can be discretized in the unitary plane and can be represented by a quaternary tree, for which it has been shown in the past that is less cumbersome than a run length code ([28]). As the algorithm is additive, several occurrences of a same domain can be accumulated at learning step so as to get a visual description rich and stable as it has already been told about moving objects in a video sequence.

Concerning the descriptors of shape rendering, it is not necessary to go beyond the quantization of their coefficients, it will be enough to discretized the expression of their coefficients over the shape domain for rebuilding the shape as it has been captured before any processing.

The description of simple shapes as it was just described may also be compared to the phonetic systems used by the present alphabets made from consonants and vowels whose conjunction allows the development of a syllabic writing: each simple shape is the conjunction of the domain of a geometric shape and an analytic expression detailing the visual rendering to be performed for rebuilding the simple shape initially captured and extracted from an image. The domain can be then assimilated to a consonant and the rendering to one or several vowels. That is why this primary visual set has be called viseme by analogy with the wording of phoneme that the syllabic writing attempt to represent: the phoneme is described as the smallest discrete unit that can be distinguished by segmentation of the speech ([3]).

## 5.3. Visual dictionary of complex shapes

Grouping simple shapes into compound shapes is at the foundation of the technique of partially hidden shape recognition proposed right here and enabling to identify tridimensional real objects observed in projective geometry by a planar sensor. According to the position and the attitude of the observer in front of the observed object, it will be produced various gathering series of simple shapes whose tree views will describe the organization and will enable to identify series of common shapes as identical branches or sub-branches. Each tree view of simple shapes describing a given compound shape consists in a word built using the previously described alphabet of visemes. So a same tridimensional object observed according to various viewpoints will generate as many words using this alphabet as they will be many occurrences for describing it: from a lexicographic viewpoint, they are each others synonymous.

So in the example of face recognition, the attribute table establishes according to this way one of the words enabling the indexing of the face that has been segmented in the aim to be labeled with the name of the person to which it is belonging. More precisely, it is only describing the consonants of its composition, the vowels being hidden by the segmentation performed in this particular application (rendering not taking in account).

The set of compound shapes found in an image then enables to build the dictionary of image shapes: it initially owns an organization similar to the one provided by a piecewise regular segmentation of the same image, but it can be distinguished by the fact that it is indexed by an alphabet of simple shapes in place of surface pieces of a given order. This dictionary may be enriched not only with the shapes belonging to a single image, but also with a video sequence by identifying the groupings that are following a common movement.

Images and video sequences can be multiplied for consolidating its building and recognizing similar sequences and identifying visual synonyms.

Therefore it is dealing with the building recognition bases by using learning:

- supervised learning by naming the observed objects in the first image of a sequence corresponding to a still video frame, then by enriching the dictionary by following still and moving objects image by image;
- unsupervised learning by leaving the enrichment process identifying by itself similar objects and by naming them afterwards the processing end.

So by labeling the similar objects of the learning base, the dictionary gains an ontological dimension in being able to name the observed objects. The richness of the vocabulary recorder into a visual dictionary varies with the depth of the indexing tree of the compound shapes that it is handling.

## *5.4. Synthesis of simple and complex shapes*

With the help of the dictionary of complex shapes located in an image, each complex shape must be decomposed into a series of simple shapes that will be put in location and place, as it has been registered:

- by placing the shapes domains there where they have been localized, that is at $(x_G, y_G, \theta)$ in the image plane after scaling with the ratio $M(u_1^2)$ ;
- by digitizing the rendering of each spectral component in the polar coordinate system of its analytic expression $(\theta_{Xu}, \theta_{XZ}, \theta_{YZ})$ centered at $\begin{pmatrix} \bar{z} \\ \bar{x} \\ \bar{y} \end{pmatrix}$ knowing that $\begin{pmatrix} \bar{x} \\ \bar{y} \end{pmatrix} = \begin{pmatrix} x_G \\ y_G \end{pmatrix}$ , and by performing a homothety with the ratio $\lambda_u$ .

For filling the holes and avoiding to have some value duplication, it should be preferable to apply a median filtering on the image of the shape domain labels before digitizing the analytic expressions of each spectral component.

# 6. Image and video coding

## *6.1. Optimal path along the blocks of a regularly divided image*

A planar image can be regularly divided into blocks of same size using treelike representations as for instance quad-trees or quaternary trees. The Hilbert or Peano-Hilbert curve enables to go all over these blocks by moving towards the nearest neighbor by crossing once and only once the same block. Having got a quad-tree model of an image, it is easy to perform a recursive traversal that provides such a curve for visiting each point of a given image ([8]).

The proposed algorithm is the following one:

- it is a recursive algorithm visiting the nodes of a quad-tree according to a depth-first traversal;
- at each tree-node it is linked an orientation that will define in which order its children nodes must be visited;
- when they are visited, the orientation of each child node is computed back from the one linked to its father so as to make the visit of its grandsons homothetic to this one used for its children more or less a given rotation;
- for different orientations can be provided each one corresponding to a building pattern of the Hilbert curve that can be deduced one from another one according to a rotation of 90°.

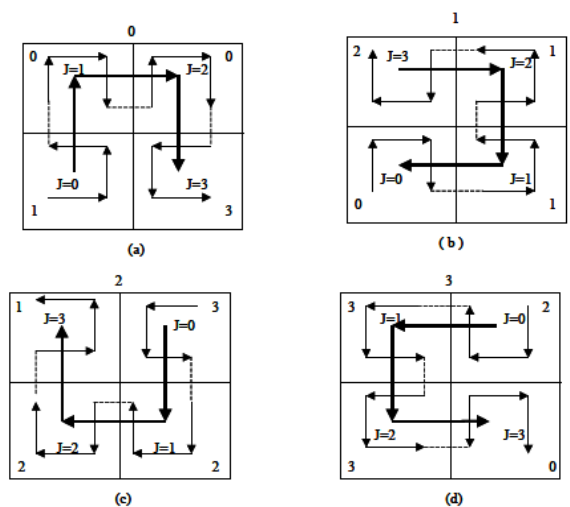

**Figure 5: Four basic patterns of a planar Hilbert curve**

These four patterns are shown above ([27]) including their developments at the following resolution.

## *6.2. Visiting path for the objects of an image*

It is proposed to use such a path for visiting the different pieces and their groupings in an image for performing its encoding. As there is an identity of structure between the image piecewise regular decomposition and the tree of the gravity centers of the same pieces, it can be attempted to visit all the pieces by moving towards the nearest neighbor according to the location of their gravity centers. That will be also equal to minimize the eye motion of an observer that would rather examine successively each shape belonging to the image by viewing them once and only once.

Here is for instance one of the possible path enabling to visit several towns in some country, by minimizing the movements from one town to another one ([29]). Their coordinates are given in a cartographic way (latitude and longitude on the terrestrial planisphere).

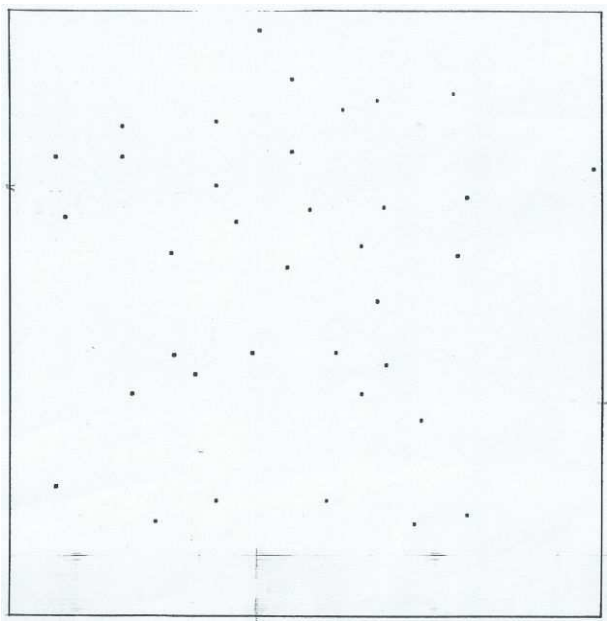

**Figure 6 : Map of the main towns of France in polar coordinates**

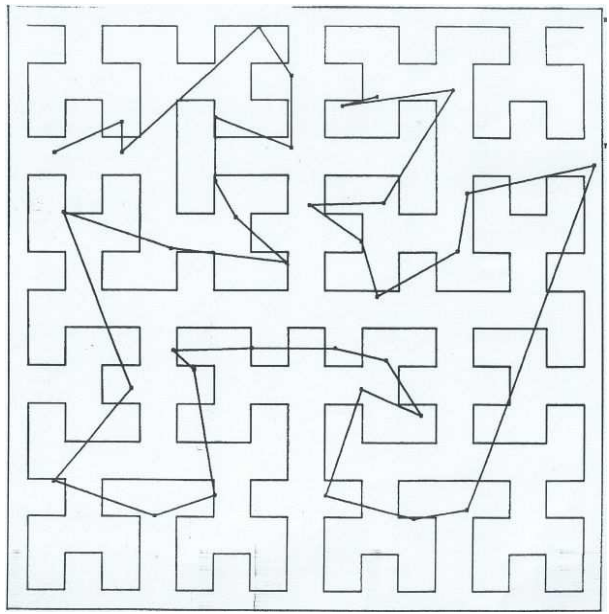

**Figure 7 : Search of a visiting path by a Hilbert scanning**

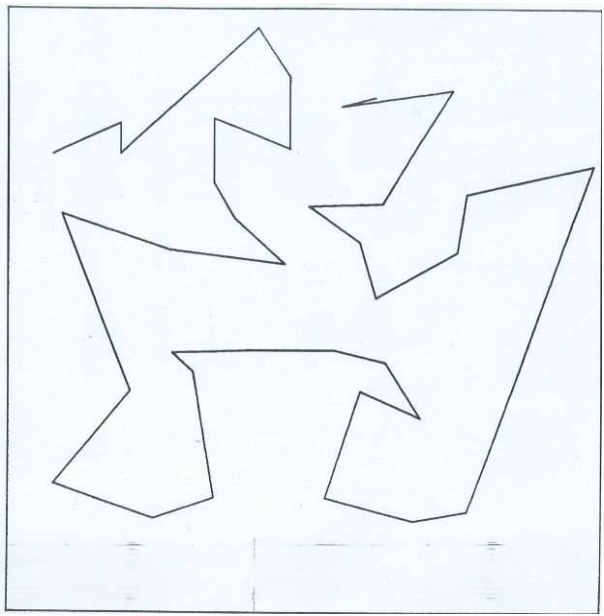

**Figure 8 : Nearest neighbor visiting path of the main towns in France**

## 6.3. Visual syntax and phraseology

Such a visiting path is envisioned for encoding the simple shapes as well as the complex shapes found in an image. It enables to list the various shapes making an image by moving from shape to shape at the nearest neighbor. From a syntactical point of view, it enables to represent an image as the sentence of a series of words registered in a dictionary of complex shapes that are themselves written using the simple shapes alphabet whose consonants are the domains of these shapes and vowels the analytic expressions of their rendering.

As the path will constitute an image scan starting nearly always at the same place and following the same motion rules, it can be envisioned that these sentences may be categorized so as to describe :

- the composition of the visual scene by referring to conventional scenes (see, land, mountain landscapes, group overview, portrait, etc.);
- the nature of the shooting (wide shot, near shot, aerial view, low-angle shot, etc.).

## 6.4. Multidimensional sorting of a shape base

The modeling of regularly decomposed spaces of dimension $k$ by trees of order $2^k$ embedded into binary trees shows that this representation model owns the cardinality of the continuum, that is to say that it can be found a continuous transformation that maps subsets from $R^k$ straight in $R$.

This means that it can be found paths enabling to visit in a continuous way once and only once the whole set of a base: in the case of $2^k$-trees by moving towards the nearest neighbor according to the Hausdorff distance.

The Hilbert curve does not only apply on planar scans, but can be also extended to multidimensional spaces by generalizing the visiting process of the quadrants of a quaternary tree to the $2^k$-ants of a $2^k$-tree.

For a multidimensional data base, it is equal to find a visiting path maximizing step by step the similarity index of registered data, that is sorting the data of the base according to their similarity.

In the research project quoted in introduction, it was envisioned to provide an image editing module where shape libraries may be built by learning and indexed using their descriptors so as to be able in reaching them in a sorted way according to their similarity in the multidimensional modeling space, moving towards the nearest neighbor at variable precision.

# Conclusion

The scientific program that has just been described enables to get nearer to the concept of generalized image such as D.H. Ballard and C.M. Brown have presented it in their book « Computer Vision » ([2]):

- the digital image is corresponding at the description level of a luminescence table;
- the segmented image enables to define homogeneous objects (segmentation of the image domain into components sharing common properties in the functional image space);
- the geometrical structure: decomposition of the segmented domains over a base of geometric primitives, description level of an image equal to graphic information;
- the relational structure: grouping of subsets of the initial image partition by discovering spatial relation or similarities with known organization models.

It echoes to the article that publishes at the same while J. Mariani in a special issue of the periodical « La Recherche » focusing on « L'Intelligence Artificielle » ([3]) and in which he describes the speech recognition as a process decomposed into four levels plus one more proper to the aimed application:

- the acoustic level: level where the phonic signal is sampled and from which are extracted the acoustic features;
- the phonetic level: from which are extracted the phonemes;
- the lexical level: where are composed words;
- the syntactic and semantic level: where are composed sentences;
- and at last the pragmatic level: level linked to the application, usable in a finite universe defined in advance, without that the previous steps cannot be validated.

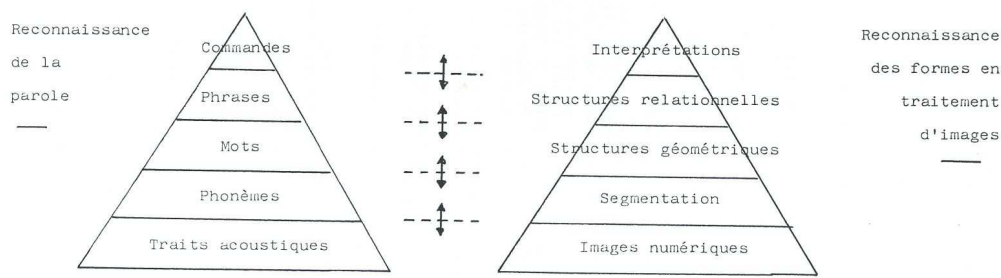

**Figure 9 : In formation pyramidal nesting in a perceptual system**

As it is shown by the developments proposed in the research program that has just been described, it cannot prevent oneself to establish a link between these two approaches of artificial perception in order to state a parallelism as it is shown above in the form of a pyramidal organized diagram.

By taking back the expected results described in the research proposal submitted to the initiative Open-FET of the VI[th] PCRD, the project SDVC ("Self-Descriptive Video Coding") tries to build the foundations of a solution proposing an innovative representation for describing the visual information that should allow to define visual ontologies and to handle visual content similarly to textual content. With the help of this modeling scheme, SDVC should provide an efficient tool for querying quickly mixed content (textual and visual) and might open a similar manner for taking in account audio contents in the future.

In this objective, the project SDVC aims to settle innovative research activities and should enable to improve the state of art about this subject:

- by proposing a description of the information at an intermediate level enabling to build more efficient retrieval systems for the image and video information;
- by taking in account the perception laws so as to bridge the gap between digitized objects and the perspective views of objects belonging to the real world;
- by managing visual objects independently to their localization in the view plan and to a set of given geometric transformations;
- by proposing an indexing scheme from which visual ontologies can be inferred and by enabling to define a kind of visual alphabet for simply connected shapes and a visual dictionary including perceptual groupings;
- and consequently by providing a means for bridging the semantic gap existing at the heart of visual information ([26]).

The project shows a strong exploratory perspective and open the path for future investigations about the representation of contents and technologies of information self-descriptive coding.

The project has been selected during the first step of the project call Open-FET (open project for future and emerging technologies) under the name of GROOVIES, but has not unfortunately be funded. It has been submitted at a while where research on multi-scale representations and geometrical wavelets was intensive ([17]-[23]). The same representations are presently mobilized in order to try and explain the efficiency of learning using multi-stage convolutional networks ([24]). The works that have just been presented perhaps might unveil parts of some questions, notably for knowing how many stages are necessary for insuring the good operation of such a network.

# Bibliographic references